\documentclass[journal]{cls/IEEEtran}

\usepackage[table,usenames,dvipsnames]{xcolor}      
\usepackage{enumitem}                               
\usepackage{extarrows}                              
\usepackage[noadjust]{cite}                         

\usepackage{amsmath,amssymb,amsfonts,amsthm,dsfont} 
\usepackage{algorithm,algorithmicx,listings}        
\usepackage[noend]{algpseudocode}			        

\usepackage{graphicx,tabularx,subcaption}
\usepackage[export]{adjustbox}
\usepackage{booktabs}
\usepackage{multirow}

\usepackage[font={small}]{caption}
\usepackage{comment}

\usepackage[titletoc,title]{appendix}

\usepackage[breaklinks=true, colorlinks, bookmarks=true, citecolor=Black, urlcolor=Violet,linkcolor=Black]{hyperref}

\def\argmin{\mathop{\arg\min}\limits}

\newcommand{\reviewera}[1]{{\color{black}#1}}
\newcommand{\reviewerb}[1]{{\color{black}#1}}
\newcommand{\reviewerc}[1]{{\color{black}#1}}
\newcommand{\reviewer}[1]{{\color{black}#1}}
\newcommand{\reviewersix}[1]{{\color{black}#1}}
\newcommand{\reviewerseven}[1]{{\color{black}#1}}
\newcommand{\reviewertwo}[1]{{\color{black}#1}}
\newcommand{\reviewerthree}[1]{{\color{black}#1}}

\newcommand{\crl}[1]{\left\{#1\right\}}

\theoremstyle{definition}

\newtheorem*{definition*}{Definition}
\newtheorem*{problem*}{Problem}

\newtheorem*{proposition*}{Proposition}

\newcommand{\calD}{{\cal D}}
\newcommand{\calE}{{\cal E}}
\newcommand{\calF}{{\cal F}}

\newcommand{\calM}{{\cal M}}

\newcommand{\calP}{{\cal P}}
\newcommand{\calQ}{{\cal Q}}

\newcommand{\bfb}{\mathbf{b}}

\newcommand{\bfp}{\mathbf{p}}
\newcommand{\bfq}{\mathbf{q}}

\newcommand{\bfv}{\mathbf{v}}

\newcommand{\bfx}{\mathbf{x}}

\newcommand{\bfalpha}{\boldsymbol{\alpha}}
\newcommand{\bfbeta}{\boldsymbol{\beta}}

\newcommand{\bftheta}{\boldsymbol{\theta}}

\newcommand{\bflambda}{\boldsymbol{\lambda}}

\newcommand{\bfrho}{\boldsymbol{\rho}}

\newcommand{\bfA}{\mathbf{A}}
\newcommand{\bfB}{\mathbf{B}}
\newcommand{\bfC}{\mathbf{C}}
\newcommand{\bfD}{\mathbf{D}}
\newcommand{\bfE}{\mathbf{E}}
\newcommand{\bfF}{\mathbf{F}}
\newcommand{\bfG}{\mathbf{G}}

\newcommand{\bfI}{\mathbf{I}}

\newcommand{\bfL}{\mathbf{L}}

\newcommand{\bfR}{\mathbf{R}}
\newcommand{\bfS}{\mathbf{S}}

\newcommand{\bfV}{\mathbf{V}}
\newcommand{\bfW}{\mathbf{W}}

\newcommand{\bbR}{\mathbb{R}}

\begin{document}
%
\title{TerrainMesh: Metric-Semantic Terrain Reconstruction from Aerial Images Using Joint 2D-3D Learning}
%
%
%

\author{Qiaojun Feng,~\IEEEmembership{Student Member,~IEEE,} and Nikolay Atanasov,~\IEEEmembership{Senior Member,~IEEE}
\thanks{We gratefully acknowledge support from NSF NRI CNS-1830399 and ARL DCIST CRA W911NF-17-2-0181.}%
\thanks{The authors are with the Department of Electrical and Computer Engineering, University of California San Diego, La Jolla, CA 92093, USA (e-mails: {\tt\footnotesize \{qjfeng,natanasov\}@ucsd.edu}).}%
}

%
%



\maketitle

\begin{abstract}
This paper considers outdoor terrain mapping using RGB images obtained from an aerial vehicle. While feature-based localization and mapping techniques deliver real-time vehicle odometry and sparse keypoint depth reconstruction, a dense model of the environment geometry and semantics (vegetation, buildings, etc.) is usually recovered offline with significant computation and storage. This paper develops a joint 2D-3D learning approach to reconstruct a local metric-semantic mesh at each camera keyframe maintained by a visual odometry algorithm. Given the estimated camera trajectory, the local meshes can be assembled into a global environment model to capture the terrain topology and semantics during online operation. A local mesh is reconstructed using an initialization and refinement stage. In the initialization stage, we estimate the mesh vertex elevation by solving a least squares problem relating the vertex barycentric coordinates to the sparse keypoint depth measurements. In the refinement stage, we associate 2D image and semantic features with the 3D mesh vertices using camera projection and apply graph convolution to refine the mesh vertex spatial coordinates and semantic features based on joint 2D and 3D supervision. Quantitative and qualitative evaluation using real aerial images show the potential of our method to support environmental monitoring and surveillance applications.
\end{abstract}

\begin{IEEEkeywords}
Mapping, Semantic Scene Understanding, Aerial Systems: Perception and Autonomy, Graph Convolution for Mesh Reconstruction
\end{IEEEkeywords}

%
\IEEEpeerreviewmaketitle

\section*{Supplementary Material}
\label{sec:supp}

\noindent See
{\url{http://erl.ucsd.edu/pages/terrainmesh.html}} for an open-source implementation, dataset, and additional results.

\section{Introduction}
\label{sec:introduction}

\begin{figure}[t]
  \centering
  \includegraphics[width=\linewidth,trim=0mm 0mm 0mm 0mm, clip]{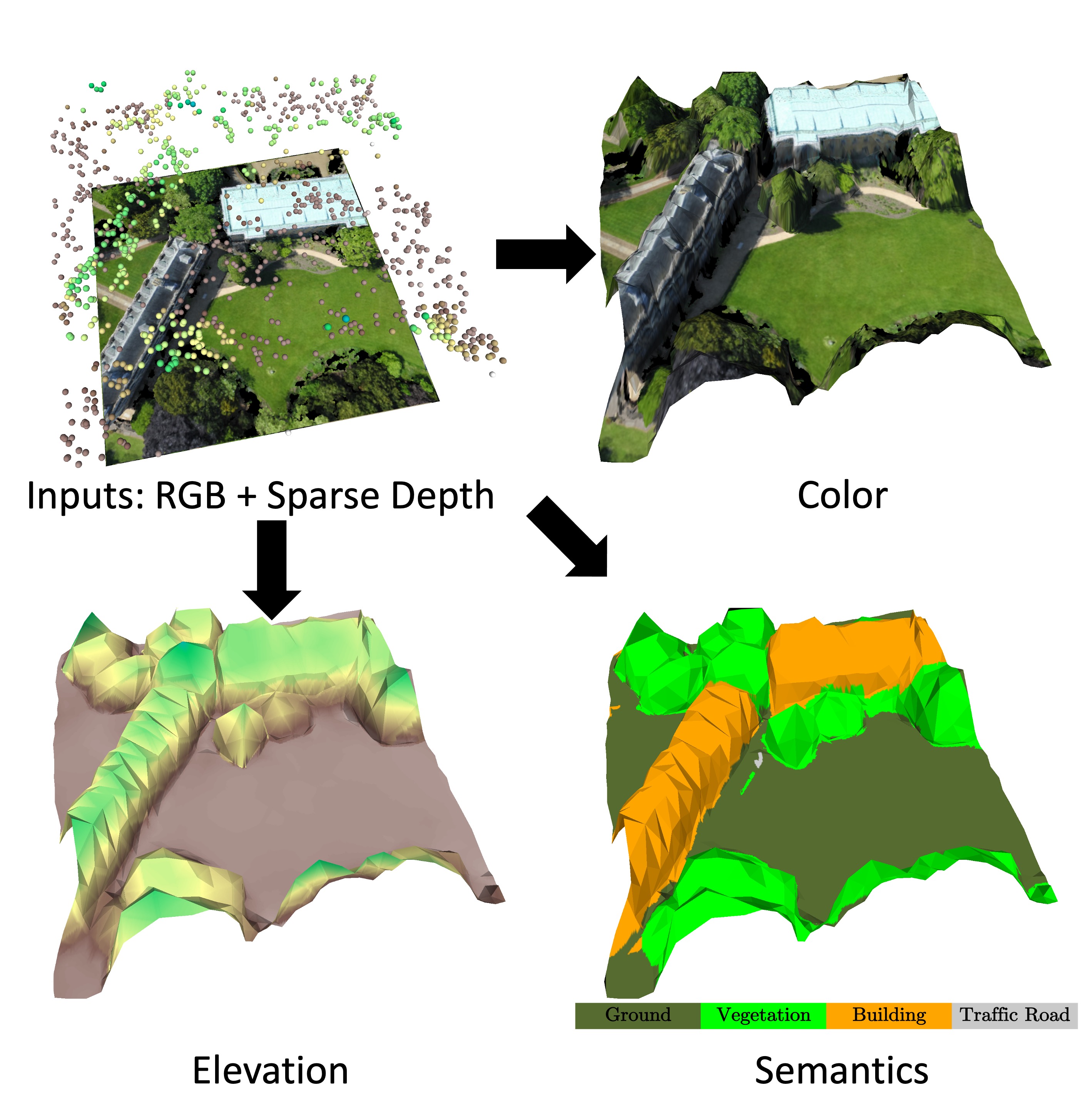}
  \caption{This paper develops a method using aerial RGB images and sparse depth measurements (top-left) to reconstruct a semantic mesh of an outdoor terrain. The color, elevation, and semantics of the mesh are visualized in the top-right, bottom-left and bottom-right plots.}
  \label{fig:intro}
\end{figure}


\IEEEPARstart{R}{ecent} advances in sensing, computation, storage and communication hardware have set the stage for mobile robot systems to impact environmental monitoring, security and surveillance, agriculture, and many other applications. Constructing terrain maps onboard an unmanned aerial vehicle (UAV) using online sensor measurements provides critical situational awareness in such applications. This paper considers the problem of building a metric-semantic terrain model, represented as a triangular mesh, of an outdoor environment using a sequence of overhead RGB images obtained onboard a UAV. Fig.~\ref{fig:intro} shows an example input and mesh reconstruction. We assume that the UAV is running a localization algorithm, based on visual-inertial odometry (VIO) \cite{Qin2018VINS} or simultaneous localization and mapping (SLAM) \cite{Cadena2016SLAM}, which estimates its camera pose and the depths of a sparse set of tracked image keypoints. However, range sensors and, hence, dense depth information are not available during outdoor flight. One approach for terrain mapping is to recover depth images at each camera view using dense stereo matching, fuse them to generate a point cloud, and triangulate a mesh surface. While specialized sensors and algorithms exist for real-time dense stereo matching, they are restricted to a limited depth range, much smaller than the distances commonly present in aerial images. Moreover, due to limited depth variation, the recovered point cloud might not be sufficiently dense for accurate mesh reconstruction. Recently, depth completion methods \cite{Ma2018Sparse,Chen2019Joint} using deep learning have shown promising performance on indoor \cite{Sturm2012RGBD} and outdoor datasets \cite{Geiger2013KITTI}. However, aerial images are different from ground RGBD images used to train these models. Due to the limited availability of aerial image datasets for supervision, learning-based methods have not yet been widely adopted for outdoor terrain mapping. Recently, there has also been increasing interest in supplementing geometric reconstruction with semantic information because many robotics tasks require semantic understanding. However, few algorithms exist for joint metric-semantic reconstruction. Most works treat semantic classification as a post-processing step, decoupling it from 3D geometric reconstruction. 




This paper is an extension of our 2021 IEEE ICRA conference paper \cite{Feng2021Mesh} on mesh reconstruction from aerial RGB images with sparse depth measurements. We propose a joint 2D-3D learning method for metric-semantic mesh reconstruction using a novel coarse-to-fine strategy, composed of mesh initialization and mesh refinement stages. In the initialization stage, we use only the sparse depth measurements to fit a coarse mesh surface. In the refinement stage, we extract deep convolutional 2D image features and associate them with the initial mesh 3D vertices through perspective projection. The mesh is subsequently refined using a graph convolution model to predict both spatial coordinates and semantic features residuals of the vertices.  We conduct extensive evaluation on simulated and real aerial datasets. The proposed mesh reconstruction method can be combined with any feature-based SLAM algorithm \cite{Campos2021ORB} to fuse local keyframe meshes into a consistent global terrain model.


\reviewera{The main contribution of the journal compared to the conference is the introduction and association of semantic segmentation with the mesh vertices, interpolation and projection techniques to obtain dense semantic features over the whole mesh, and graph convolution network optimization with a joint geometric-semantic loss function to optimize the mesh vertices and semantic features. We demonstrate empirically that the joint geometric-semantic training can outperform the earlier geometric-only method proposed in the conference paper. We also derive a closed-form mesh initialization method, which is more accurate and efficient than the one in the conference paper}, as well as an explicit mesh merging method to combine multi-view mesh reconstruction into a single consistent global mesh of the environment. In summary, the \textbf{contributions} of this paper are summarized as follows.
\begin{itemize}
    \item We introduce a joint 2D-3D loss function, utilizing differentiable mesh rendering, for metric-semantic mesh reconstruction.
    \item We develop a two-stage coarse-to-fine mesh reconstruction approach, using a closed-form mesh vertex \emph{initialization} from sparse depth measurements and a graph convolution network mesh vertex \emph{refinement} from RGB, sparse depth measurements and semantic image features.
    \item We evaluate our metric-semantic mesh reconstruction algorithm on synthetic and photo-realistic aerial image datasets. 
\end{itemize}




\section{Related Work}
\label{sec:related_work}

\subsection{Depth Completion}
Predicting depth from monocular RGB images allows single-camera perception systems to recover 3D environment structure \cite{Godard2019Monodepth2,Gordon2019Depth}. However, dense depth estimation from monocular images may be challenging, especially for aerial images, where the depth variation is small compared to the absolute depth values. In contrast, the depth of sparse visual keypoints may be obtained efficiently and accurately using triangulation \cite{Hartley2004} between tracked feature points from Kanade-Lucas-Tomasi tracking \cite{Shi1994Good} or visual feature matching \cite{Rublee2011ORB}.

\reviewerseven{
Depth completion is the task of reconstructing a dense depth image from an RGB image with given sparse depth estimates. Ma et al. \cite{Ma2018Sparse,Ma2019SelfSupervised} develop a deep network for depth completion that passes the sparse depth and RGB image inputs through convolution layers, ResNet encoder layers, transposed convolution decoder layers, and a 1x1 convolution filter. The model is trained either with supervision from ground-truth depth images or via photometric error self-supervision from calibrated RGB image pairs. Instead of consuming sparse depth images directly, Chen et al. \cite{Chen2018Depth} pre-process sparse depth images by generating a Euclidean distance transform of the keypoint locations and a nearest-neighbor depth fill map. The authors propose a multi-scale deep network that treats depth completion as residual prediction with respect to the nearest-neighbor depth fill maps. 
We borrow the idea of densify the 2D sparse depth inputs for our model design.
Chen et al. \cite{Chen2019Joint} design a 2D convolution branch to process stacked RGB and sparse depth images and a 3D convolution branch to process point clouds and fuse the outputs of the two branches so that the 2D and 3D features are combined. 
Our approach also combines both the 2D and 3D feature extraction phases but in the form of mesh.
CodeVIO \cite{Zuo2021CodeVIO} uses a Conditional Variational Autoencoder (CVAE) to encode RGB and sparse depth inputs into a latent depth code and decode a dense depth image from the latent depth code. The sparse depth measurements are used to perform incremental depth code updates, allowing the depth reconstruction to be coupled with visual odometry estimation in the MSCKF filter \cite{Mourikis2007MSCKF}. Our method is loosely integrated with a feature-based VIO/SLAM algorithm.
}

We approach depth completion using mesh reconstruction from RGB images and sparse depth measurements. While both dense depth completion and mesh reconstruction are challenging, a mesh model is more memory efficient than a depth image. For example, a $512\times 512$ dense depth image requires $260,000$ parameters. In contrast, our approach can represent the same region using only 1024 vertices.


\subsection{Mesh Reconstruction}
\reviewerthree{
Online terrain mapping requires efficient storage and updates of a 3-D surface model. However, storing dense depth information from aerial images needs significant memory and subsequent model reconstruction for robot objectives like motion planning or environment exploration. An explicit surface representation using a polygonal mesh can be quite memory and computationally efficient. Compared to sparse point cloud models, mesh surfaces are continuous, allowing direct integration in motion planning algorithms as well as intuitive visualization for human operators.}
FLaME \cite{Greene2017FlaME} performs variational optimization over a time-varying Delaunay graph to obtain an inverse-depth mesh of the environment, using sparse depth measurements from a VIO algorithm.
Rosinol et al. \cite{Rosinol2021Smooth} extend FLaME to optimize the mesh over dense depth image measurements in real-time using a parallel implementation. Rosinol et al. \cite{Rosinol2019Mesh} detect vertical and horizontal planes to regularize mesh vertices, and optimize the mesh vertices and the camera poses using a factor graph. Voxblox \cite{Oleynikova2017Voxblox} incrementally builds a voxel-based truncated signed distance field and can reconstruct a mesh as a post-processing step using the Marching Cubes algorithm \cite{Lorensen1987Marching}. 
Terrain Fusion \cite{Wang2019TerrainFusion} performs real-time terrain mapping by generating digital surface model (DSM) meshes at selected keyframes. The local meshes are converted into grid-maps and merged using multi-band fusion. 

Recently, learning methods have emerged as a promising approach for mesh reconstruction from limited or no 3D information. Bloesch et al.~\cite{Bloesch2019Mesh} propose a learning method to regress the image coordinates and depth of mesh vertices in a decoupled manner. This allows an in-plane 2D mesh to capture the image structure. 
Pixel2Mesh \cite{Wang2018Pixel2Mesh} treats a mesh as a graph and applies graph convolution \cite{kipf2017semi} for vertex feature extraction and graph unpooling to subdivide the mesh for refinement. Using differentiable mesh rendering \cite{Kato2018Neural,Liu2020Softras}, the 3D mesh structure of an object can be learned from 2D images \cite{Kanazawa2018ECCV,Feng2019Mesh,Simoni2021Multi}. Mesh R-CNN \cite{Gkioxari2019MeshRCNN} simultaneously detects objects and reconstructs their 3D mesh shape. A coarse voxel representation is predicted first and then converted into a mesh for refinement. Recent works \cite{Nie20203D,Weng2021Holistic} can generate mesh reconstructions of complete scenes, including object and human meshes and their poses, from a single RGB image. 

In contrast with many mesh reconstruction approaches, our method uses both visual and semantic features to refine the mesh geometry and generates mesh models with per-vertex semantic category distributions.


\subsection{Semantic 3D Reconstruction}

3D reconstruction from an image sequence is a fundamental problem in robotics and computer vision. Multi-view stereo (MVS) \cite{Furukawa2015MVS} aims to estimate the depth at one frame using several different frames. Classical MVS methods perform patch matching with photometric and geometric consistency \cite{schoenberger2016mvs}. These methods generalize well although the performance can be affected by low texture, lighting variation, and occlusion. 
Recently, learning-based methods that fuse multi-view learned features across frames for depth recovery have achieved excellent performance. NeuralRecon \cite{Sun_2021_CVPR} reconstructs and fuses sparse TSDF volumes for each frame incrementally using 3D sparse convolutions and Gated Recurrent Units (GRUs). VoRTX \cite{Stier2021VoRTX} learns to fuse multi-view frames using a transformer model and projective occupancy. SimpleRecon \cite{Sayed2022SimpleRecon} leverages relative poses among frames to build a cost volume and uses a multi-layer perceptron (MLP) to reduce the volume and avoid costly 3D convolution. Learning-based MVS \cite{Murez2020Atlas,Arda2021DeepVideoMVS,Sun2022NeuconW,FineRecon} can tackle challenges like severe occlusion but generally labeled data is needed for training. Recently, SLAM systems that integrate a learning-based MVS model to obtain dense 3D reconstruction in real-time have been proposed \cite{Koestler2022TANDEM,Xin2023SimpleMapping}.
Besides explicit 3D representations, implicit representations that model 3D structures as level sets of distance or radiance functions have shown impressive performance recently. 
\reviewertwo{Neural Radiance Fields (NeRF) \cite{Mildenhall2020NeRF} represent the color and density field of the scene through the weights of the neural network. For any given camera pose, an RGB and depth image can be generated through volume rendering over the NeRF, providing photo-realistic novel view synthesis. Another advantage of NeRF models is that they can be trained from posed RGB images without depth supervision. However, for any new 3D scene a NeRF needs to be trained from scratch which may take a long time. Numerous recent works have extended the NeRF model \cite{zhang2022nerfusion,Barron2022MipNeRF360,Wang2022HFNeuS,Li2023Neuralangelo} to enable MVS pre-training, capture high-frequency details, and apply to unbounded outdoor environments. Instant-NGP \cite{Muller2022NGP} and Nerfstudio \cite{Tancik2023Nerfstudio} incorporate many of the recent NeRF model advances and offer open-source implementations.}


\reviewerthree{
Semantic 3D reconstruction aims to estimate both the geometric structure and semantic content of an environment from visual observations. Extending 2D semantic segmentation and depth prediction to a 3D multi-view consistent semantic model provides information that is critical for environmental monitoring tasks, such as observing vegetation recovery after a wildfire or controlling fuel build-up for fire prevention \cite{Samiappan2019Wildfire}, or for terrain traversablility estimation in ground-aerial robot teaming \cite{Miller2022Together}. Semantic information also allows human operators to specify tasks for mobile robots in terms of objects and concepts in the environment model.}
Semantic segmentation on the 2D images can be back-projected onto 3D space and multi-view information can be fused to annotate the 3D structure \cite{Valentin2013Mesh,Rosinol2020Kimera,Wei2020Semantic}. Besides, semantic segmentation can be directly performed on the 3D point cloud \cite{Hu2021Sensat,wang2022rangeudf} or the mesh \cite{Hu2022Voxel}. Instead of treating the semantic annotation as the post-processing step after geometric reconstruction, researchers also investigate on how to jointly optimize geometric and semantic accuracy. H{\"a}ne et al. \cite{Hane2017Dense} formulate a joint segmentation and dense reconstruction problem on voxels and show that appearance likelihoods and class-specific geometric priors help each other. Cherabier et al. \cite{Cherabier2018Priors} leverage variational energy minimization method for regularization to capture complex dependencies between the semantic labels and the 3D geometry. Guo et al. \cite{Guo2022Neural} jointly optimize the geometry and semantics by predicting the implicit neural representations of the signed distance, color and semantic field.

Our approach is most closely related to the concurrent work on semantic mesh mapping \cite{Herb2021Lightweight}. Our formulation utilizes sparse depths obtained from keyframe-based SLAM and emphasizes the interaction of geometric reconstruction and semantic segmentation in 3D mesh reconstruction.

\section{Problem Formulation}
\label{sec:problem_formulation}

Consider a UAV equipped with an RGB camera flying outdoor. Let $\bfI$ denote an RGB image. Obtaining dense depth images during outdoor flight is challenging due to the large distances and relative small variation. However, a VIO or SLAM algorithm can track and estimate the depth of a sparse set of image feature points. \reviewerb{Let $\bfD^s$ be a \emph{sparse} 2D matrix that contains estimated depths at the image feature locations and zeros everywhere else.} Let $\bfD$ denote the \emph{dense} ground-truth depth image. Let $\bfS$ denote an associated ground-truth semantic segmentation image. \reviewerseven{Assuming there are $s$ semantic classes in total, we model $\bfS$ as a tensor with the same width and height as the RGB image $\bfI$ and third channel size $s$.} Each element $\bfS_{i,j} \in [0,1]^s$ is a one-hot vector with $0$s in all elements, except for a single $1$ indicating the true semantic class.

\reviewerb{Our goal is to construct an explicit model of the camera view using a 3D semantic triangle mesh $\calM := (\bfV,\bfC,\calE,\calF)$, where $\bfV \in \mathbb{R}^{n \times 3}$ are the vertex spatial coordinates, $\bfC \in \mathbb{R}^{n \times s}$ are the vertex semantic features, $[n] := \crl{1,\ldots,n}$ is the set of vertex indices, $\calE \subseteq [n] \times [n]$ are the edges, and $\calF \subseteq [n] \times [n] \times [n]$ are the faces. Each row of the matrix $\bfC$ contains an unnormalized score vector for the $s$ classes that can be converted into a probability distribution over the $s$ classes using the softmax function \cite{Bridle1989Softmax}.}

\begin{problem*}
Given a finite set of RGB images $\crl{\bfI_k}_k$ and corresponding sparse depth measurements $\crl{\bfD^s_k}_k$, define a semantic mesh reconstruction function $\calM = f(\bfI, \bfD^s ; \bftheta)$ and optimize its parameters $\bftheta$ to fit the ground-truth depth $\crl{\bfD_k}_k$ and semantic segmentation $\crl{\bfS_k}_k$ images:
\begin{equation}
\label{eq:problem}
\min_{\bftheta} \sum_k \ell( f(\bfI_k, \bfD^s_k ; \bftheta); \bfD_k, \bfS_k )
\end{equation}
where $\ell(\calM;\bfD,\bfS)$ is a loss function measuring the error between a 3D semantic mesh $\calM$ and a depth image $\bfD$ plus a semantic image $\bfS$. 
\end{problem*}

The choice of loss function $\ell$ is discussed in Sec. \ref{sec:loss}. We develop a machine learning approach consisting of an offline training phase and an online mesh reconstruction phase. During training, the parameters $\bftheta$ are optimized using a training set $\calD := \crl{ \bfI_k, \bfD^s_k, \bfD_k, \bfS_k }_k$ with known ground-truth depth images and semantic segmentation images. During testing, given streaming RGB images $\bfI$ and sparse depth measurements $\bfD^s$, 
\reviewerseven{the optimized parameters $\bftheta^*$ are used in the model $f(\bfI, \bfD^s; \bftheta^*)$ to reconstruct the mesh vertex spatial coordinates $\bfV$ and semantic features $\bfC$. The mesh edges $\calE$ and faces $\calF$ are assumed fixed and known, and hence are not reconstructed by the model. For notational simplicity, we write the output of model $f$ directly as the semantic mesh $\calM = f(\bfI, \bfD^s; \bftheta^*)$.}
\reviewerb{A keyframe-based VIO or SLAM algorithm estimates the positions $\bfp$ and orientations $\bfR$ of camera keyframes as well as the depth of sparse keypoint measurements associated with each keyframe. Our approach estimates a local mesh $\calM = (\bfV,\bfC,\calE,\calF)$ at each camera keyframe. \reviewersix{The keyframe meshes can be converted to a global frame (with vertex coordinates $\bfV \bfR^\top + \mathbf{1} \bfp^\top\!$, where $\mathbf{1}$ is a $n\times 1$ vector filled with 1) and fused to obtain a complete consistent metric-semantic model of the environment.}}


\begin{figure}[t]
  \centering
  \includegraphics[width=\linewidth,trim=0mm 0mm 0mm 0mm, clip]{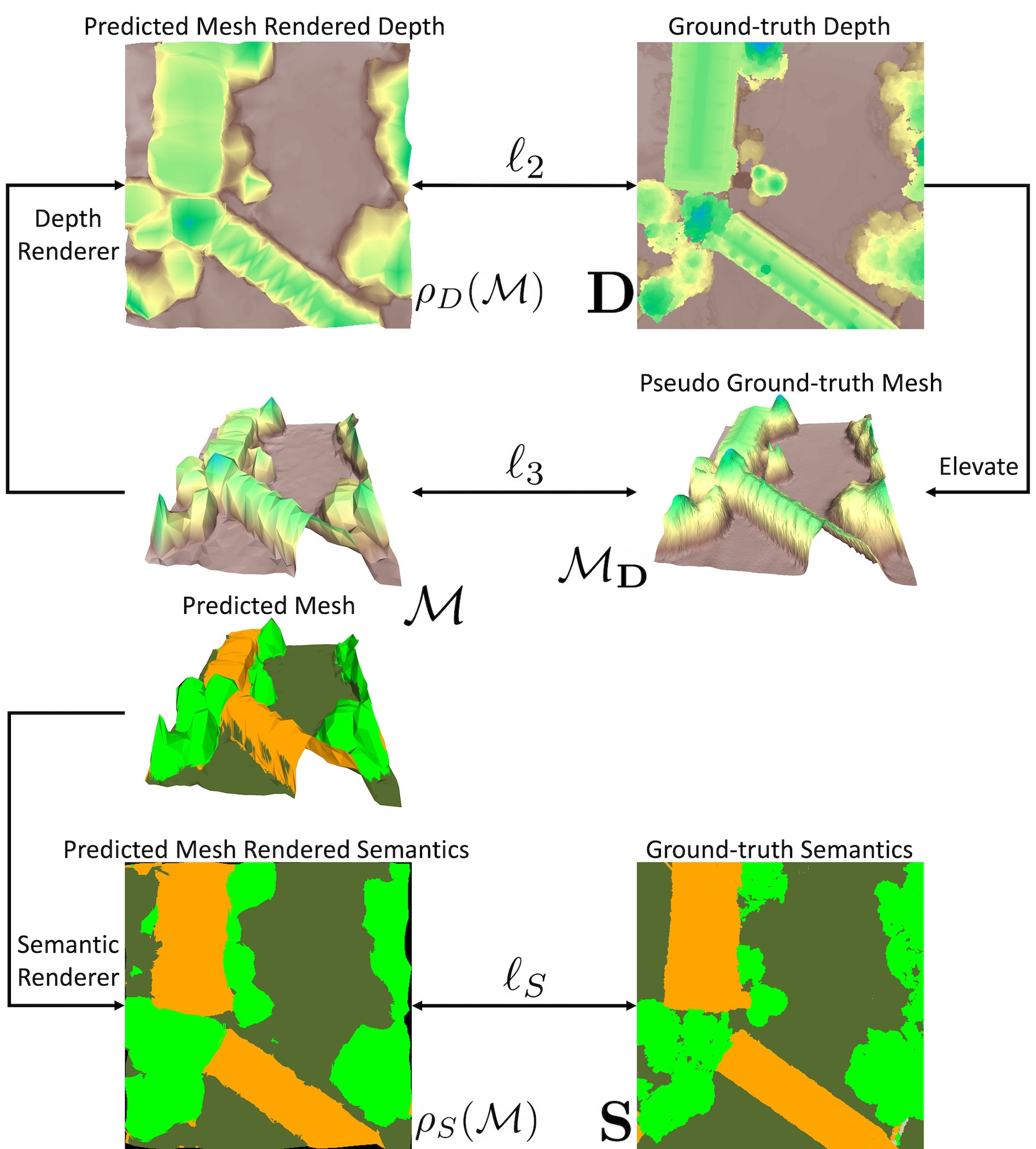}
  \caption{Loss function visualization: $\ell_{2}$ compares rendered mesh depth $\rho_D(\calM)$ to a depth image $\bfD$, $\ell_{3}$ compares a mesh $\calM$ to an elevated mesh $\calM_{\bfD}$ obtained from a depth image, and $\ell_{S}$ compares a rendered mesh semantic image $\rho_S(\calM)$ to a semantic segmentation image $\bfS$.}
  \label{fig:supervision}
\end{figure}

\begin{figure*}[t]
  \centering
  \includegraphics[width=\linewidth,trim=0mm 0mm 0mm 0mm, clip]{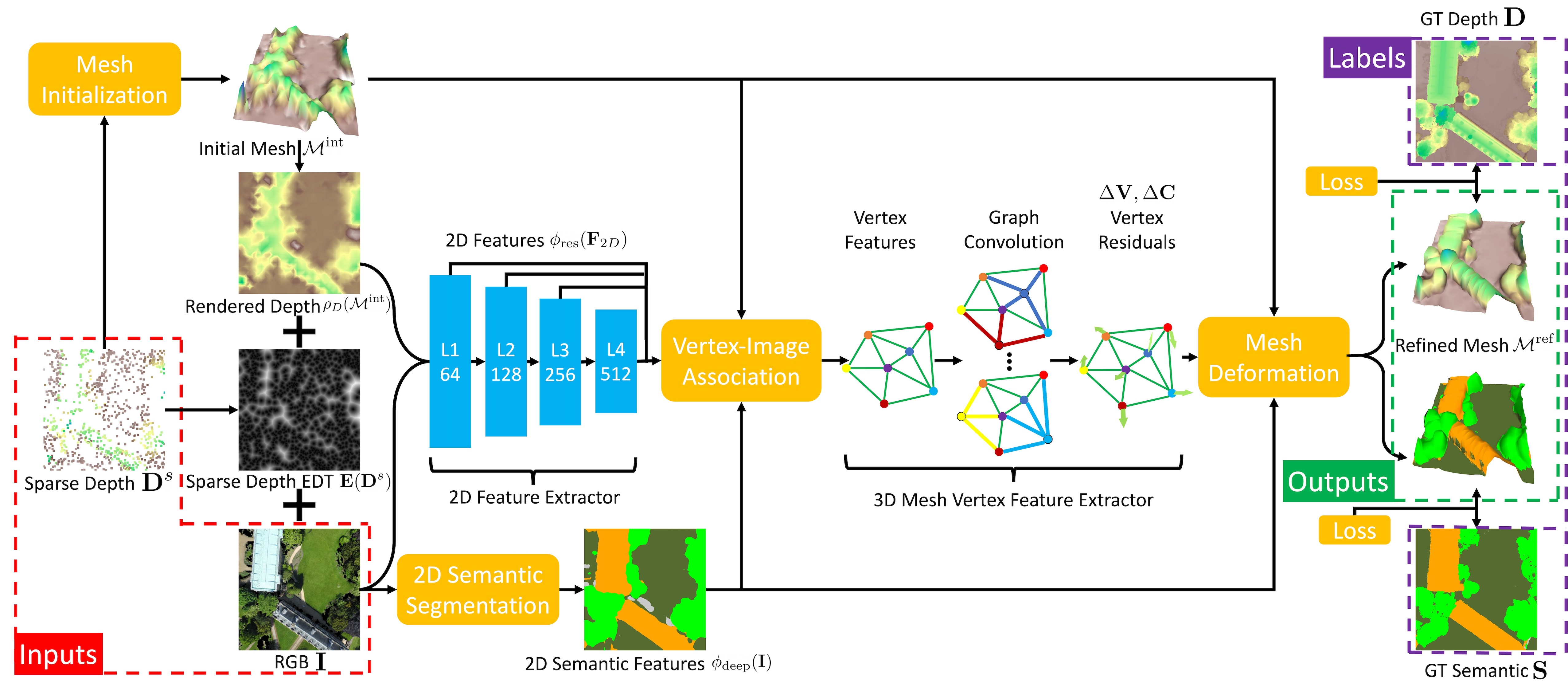}
  \caption{Overview of our mesh reconstruction architecture. In the initialization stage \reviewerc{(Sec.~\ref{sec:mesh_init})}, we use sparse depth to elevate a flat mesh from the image plane to 3D space (Fig. \ref{fig:mesh_init}). In the refinement stage (Sec.~\ref{sec:geo_mesh_refine},\ref{sec:semantic}), we first combine the RGB image, a depth image rendered from the initial mesh, and a Euclidean Distance Transform of the sparse depth measurements to extract features using a 2D feature extractor. We have a 2D semantic segmentation model to generate 2D semantic features from the RGB image. The 2D features and the 2D semantic features are \reviewerb{associated with} the mesh vertices using camera projection at different stages (Fig. \ref{fig:vertex2image},\ref{fig:mesh_sem_init}). The vertex spatial coordinates and the vertex semantic features, are regressed using graph convolution network (GCN) over the mesh. The refined output is a metric-semantic mesh (Fig. \ref{fig:refinement}). The 2D feature extractor and GCN parameters are optimized jointly using the loss function in Sec.~\ref{sec:loss}. The 2D semantic segmentation model is trained separately.}
  \label{fig:pipeline}
\end{figure*}

\section{Loss Functions for Mesh Reconstruction}
\label{sec:loss}

We develop several loss functions to measure the consistency between a semantic mesh $\calM$ and corresponding depth image $\bfD$ and semantic segmentation image $\bfS$. Since our problem focuses on optimizing the mesh, the loss function must be differentiable with respect to the mesh vertex spatial coordinates $\bfV$ and semantic features $\bfC$. We keep the mesh edges $\calE$ and faces $\calF$ fixed during the mesh optimization. 

A loss function can be defined in the 2D image plane by rendering a depth image from $\calM$ and comparing it with $\bfD$. The differentiable mesh renderer \cite{Liu2020Softras,ravi2020pytorch3d} makes the 3D mesh rendering, e.g., from a 3D mesh to a 2D image, differentiable. Therefore, we can back-propagate the loss measured on the 2D images to the 3D mesh vertices. We leverage a differentiable mesh renderer to generate a depth image $\rho_D(\calM)$ and define a 2D loss function:
\begin{equation}
  \label{eq:loss_2D}
  \ell_{2}(\mathcal{M},\bfD) := \text{mean}(|\rho_{D}(\mathcal{M}) - \bfD|),
\end{equation}
where $\text{mean}(\cdot)$ is a function taking the mean over all the valid pixels where both $\bfD$ and $\rho_{D}(\calM)$ have a depth value.

While $\ell_2$ is a natural choice of a loss function in the image plane, it does not emphasize two important properties for mesh reconstruction. First, since $\ell_2$ only considers a region in the image plane where both depth images have valid information, its minimization over $\calM$ may encourage the mesh $\calM$ to shrink to cover only a smaller image region.
Second, $\ell_2$ does not emphasize regions of large depth gradient variation (e.g., the side surface of a building), which may lead to inaccurate 3D reconstruction. To address these limitations, we define an additional loss function in the 3D spatial domain using two point clouds $\calP_{\calM}$ and $\calQ_{\bfD}$ obtained from $\calM$ and $\bfD$, respectively:
\begin{equation}
  \label{eq:loss_3D}
  \ell_{3}(\mathcal{M},\bfD) := \frac{1}{2} d(\calP_{\calM},\calQ_{\bfD}) + \frac{1}{2}d(\calQ_{\bfD},\calP_{\calM}),
\end{equation}
\reviewerseven{where $d$ is the asymmetric Chamfer point cloud distance \cite{Barrow1977Chamfer}:
%
\begin{equation}
d( \calP, \calQ) := \frac{1}{|\calP|} \sum_{\bfp \in \calP} \min_{\bfq \in \calQ}\|\bfp - \bfq\|_2^2.
\end{equation}
}
\reviewer{Note that squared Euclidean distance is used when calculating the Chamfer distance.} To generate $\calP_{\calM}$, we sample on the faces of $\calM$ uniformly using PyTorch3D library \cite{ravi2020pytorch3d}. The loss function is differentiable with respect to the mesh vertices because the samples on the mesh faces can be represented as linear combinations of the mesh vertices using the barycentric coordinate introduced in Sec. \ref{sec:mesh_init}. To generate $\calQ_{\bfD}$, we may sample the depth image $\bfD$ uniformly and project the samples to 3D space but this will not generate sufficient samples in the regions of large depth gradient variation. Instead, we first generate a pseudo ground-truth mesh $\calM_{\bfD}$ by densely sampling pixel locations in $\bfD$ as the mesh vertices and triangulating on the image plane to generate faces. We then sample the surface of $\calM_{\bfD}$ uniformly to obtain $\calQ_{\bfD}$.
\reviewersix{The sample number is set as 10000.}

We also define two regularization terms to measure the smoothness of the mesh $\calM$. The first is based on the Laplacian matrix $\bfL := \bfG - \bfA \in \mathbb{R}^{n \times n}$ of $\calM$, where $\bfG$ is the vertex degree matrix and $\bfA$ is the adjacency matrix. We define a vertex regularization term based on the $\ell_{2,1}$-norm~\cite{Nie2010efficient} of the degree-normalized Laplacian \cite{Sorkine2004Laplacian} \reviewerseven{$\bfL_{n} = \bfG^{-1}\bfL = \bfI_n - \bfG^{-1}\bf{A}$ where $\bfI_n$ is an identity matrix of size $n \times n$:}
\begin{equation}
\label{eq:loss_laplacian}
\ell_{\bfV}(\calM) := \frac{1}{n} \left\|\bfL_{n}\bfV\right\|_{2,1},
\end{equation}
where $n$ is the number of vertices. We also introduce a mesh edge regularization term to discourage long edges in the mesh
\begin{equation}
\label{eq:loss_edge}
\ell_{\calE}(\calM) := \frac{1}{|\calE|} \sum_{(i,j) \in \calE} \|\bfv_i - \bfv_j\|_2,
\end{equation}
where $\bfv_i\in\mathbb{R}^3$ are the coordinates of the $i$-th mesh vertex.

We also define a semantic loss function that relates the 3D mesh semantic information to the 2D semantic segmentation image by rendering the semantic mesh similar to (\ref{eq:loss_2D}). We define a differentiable semantic rendering function $\rho_S(\calM)$ which can generate a same-sized image as $\bfS$ with $s$ channels, where $s$ is the number of semantic classes. At each pixel, the $s$-dimensional vector stores the unnormalized scores representing the likelihoods of the $s$ classes. We use a softmax function \cite{Bridle1989Softmax} $\sigma_i(\bfx) = \exp(\bfx_i)/\sum_{j=1}^s \exp(\bfx_j)$ to compute the probability distribution over the $s$ classes $\sigma(\rho_S(\calM))$. For the semantic segmentation task, we choose the Dice loss \cite{Dice1945} :
\begin{equation}
\label{eq:loss_sem_dice}
\ell_{\bfS}(\calM,\bfS) := - \frac{2 |\sigma(\rho_{S}(\calM)) \cdot \bfS|}{|\sigma(\rho_{S}(\calM))| + |\bfS|},
\end{equation}
where $|\cdot|$ sums up all the absolute values of the elements. Note that $\bfS$ contains one-hot vectors while $\sigma(\rho_{S}(\calM))$ stores probability vectors for the $s$ classes. 
Therefore, $|\sigma(\rho_{S}(\calM)) \cdot \bfS|$ is the probabilistic intersection between two semantic segmentation images. \reviewerc{In Sec.~\ref{sec:ablation}, we compare the Dice loss with three alternative semantic loss functions (cross-entropy loss, focal loss, and Jaccard loss).} Finally, we apply Laplacian smoothing \eqref{eq:loss_laplacian} to the vertex semantic features:
\begin{equation}
\label{eq:loss_laplacian_sem}
\ell_{\bfC}(\calM) := \frac{1}{n} \left\|\bfL_{n}\bfC\right\|_{2,1}.
\end{equation}

The complete loss function is:
\begin{equation}
\label{eq:final_loss}
\begin{aligned}
\ell(\calM, \bfD, \bfS) &:= w_2 \ell_2(\calM, \bfD) + w_3 \ell_3(\calM, \bfD)\\
&\quad + w_{\bfV} \ell_{\bfV}(\calM) + w_{\calE} \ell_{\calE}(\calM)\\
&\quad + w_{\bfS}\ell_{\bfS}(\calM,\bfS) + w_{\bfC}\ell_{\bfC}(\calM)
\end{aligned}
\end{equation}
where the first two terms evaluate the error between $\calM$ and $\bfD$, the following two terms encourage smoothness of the mesh structure, and the last two terms evaluate the error between $\calM$ and $\bfS$ and regularize the semantic features, which affects both the geometric and semantic properties of the mesh. The scalars $w_2, w_3, w_{\bfV}, w_{\calE}, w_{\bfS},w_{\bfC} \in \bbR_{\geq 0}$ allow appropriate weighting of the different terms in \eqref{eq:final_loss}. Fig.~\ref{fig:supervision} illustrates the loss functions $\ell_2$ in \eqref{eq:loss_2D}, $\ell_3$ in \eqref{eq:loss_3D}, and $\ell_S$ in \eqref{eq:loss_sem_dice}. 

\section{2D-3D Learning for Semantic Mesh Reconstruction}
\label{sec:technical_approach}


Inspired by depth completion techniques, we approach mesh reconstruction in two stages: \emph{initialization} and \emph{refinement}. In the initialization stage, we generate a mesh from the sparse depth measurements alone (Sec. \ref{sec:mesh_init}). In the refinement stage, we optimize the mesh vertex coordinates based on RGB image features (Sec. \ref{sec:geo_mesh_refine}) and assign semantic categories to each vertex using image segmentation features (Sec. \ref{sec:semantic}). An overview of our semantic mesh reconstruction model $\calM = f(\bfI,\bfD^s;\bftheta)$ is shown in Fig. \ref{fig:pipeline}.

\subsection{Mesh Initialization}
\label{sec:mesh_init}

\begin{figure}[t]
  \centering
  \includegraphics[width=\linewidth,trim=0mm 0mm 0mm 0mm, clip]{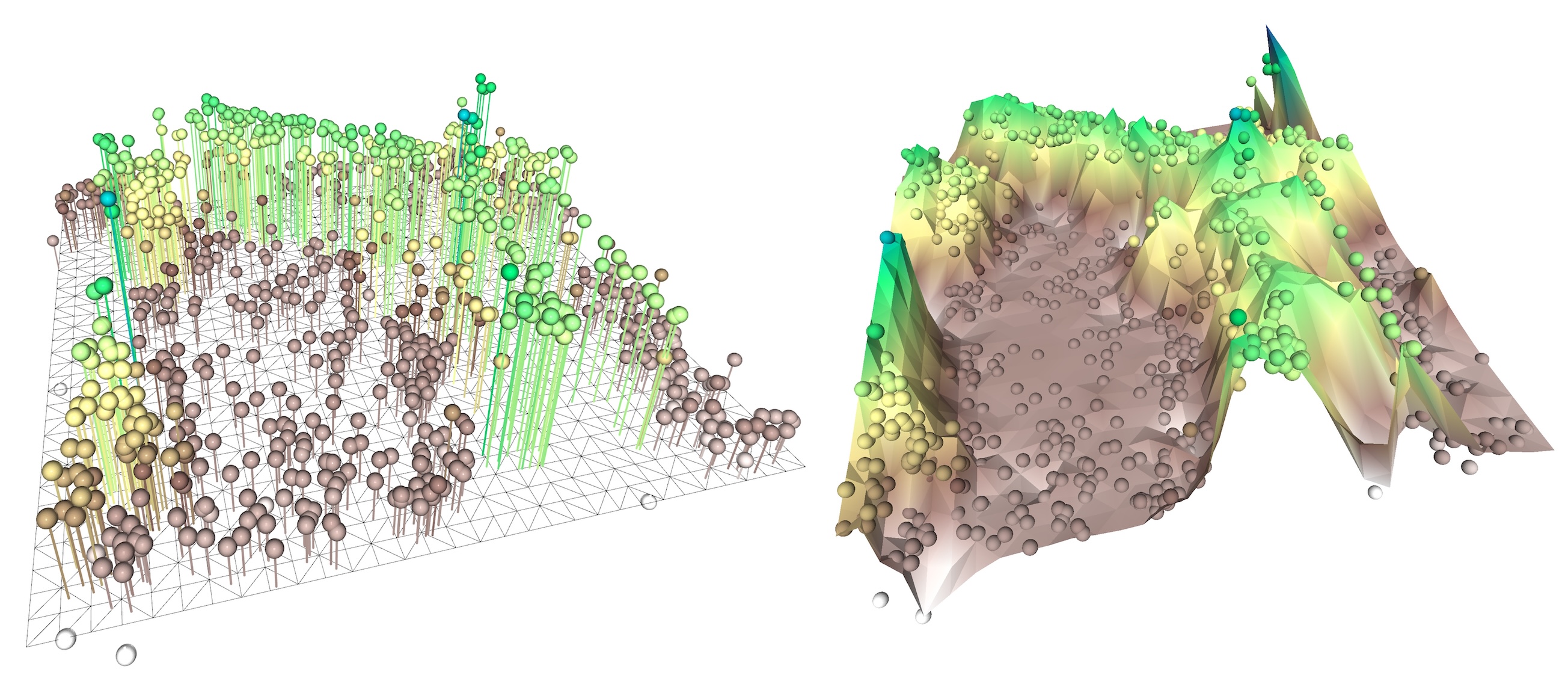}
  \caption{Mesh initialization stage \reviewerc{(Sec.~\ref{sec:mesh_init})}. Left: Sparse depth measurements (color dots) are used to determine vertex heights from a flat image-plane mesh (bottom wireframe). Right: The initialized mesh. Colors indicate elevation.}
  \label{fig:mesh_init}
\end{figure}

Outdoor terrain structure can be viewed as a 2.5-D surface with height variation. Hence, we initialize a flat mesh surface and change the surface elevation based on the sparse depth measurements. 
\reviewerc{The flat mesh is initialized with regular-grid vertices ($n=1024$ in our experiments) over the image plane, and the edges and the faces connecting the vertices.}
See Fig.~\ref{fig:mesh_init} for an illustration. Subsequently, our mesh reconstruction approach only optimizes the mesh vertices and keeps the edge and face topology fixed. 
The initialized mesh $\mathcal{M}^{\text{int}} = (\bfV ^*, \mathbf{0})$ is used as an input to the mesh refinement stage, described in Sec.~\ref{sec:geo_mesh_refine}, \ref{sec:semantic}. Since we do not update $\calE, \calF$, we will omit them for simplicity.

\reviewerb{
We constrain the mesh vertex deformation to the $z$-axis to change the vertex heights only. The coordinates of the $i$-th vertex of the flat mesh, $\bfv_i = [v^x_i,v^y_i,1]$, are divided by a scalar inverse depth $\lambda_i$ to obtain the $i$-th vertex coordinates $[v^x_i/ \lambda_i, v^y_i/\lambda_i, 1/\lambda_i]$ of the initialized mesh. We concatenate $\lambda_i$ to obtain a vector $\bflambda \in \mathbb{R}^{n}$ of all vertex inverse depths. 

Any point $\bfp$ on the mesh surface that lies in a specific triangle can be represented as a convex combination $\bfp = b_i\bfv_i + b_j\bfv_j + b_k\bfv_k$ of the triangle vertices $\bfv_i,\bfv_j,\bfv_k$ with weights $b_i, b_j, b_k\in [0,1]$ such that $b_i+b_j+b_k = 1$. The vector $[b_i,b_j,b_k]^\top$ is called the \emph{barycentric coordinates} of $\bfp$. 
We use barycentric coordinates to relate the sparse depth measurements $\bfD^s$ to the vertex inverse depths $\bflambda$, which is equivalent to a linear interpolation. 

Let the valid measurements in the sparse depth image $\bfD^s$ be $\{(i,j),\bfD^s_{ij}\}$, where $(i,j)$ are the pixel coordinates and $\bfD^s_{ij}$ are the corresponding depth measurements. Each pixel $(i,j)$ falls within one triangle of the flat 2D mesh (see Fig. \ref{fig:mesh_init}). Let $\bfb_{ij} \in \bbR^n$ be the barycentric coordinates of pixel $(i,j)$, where at most three elements of $\bfb_{ij}$, corresponding to the three triangle vertices, are non-zero. The inverse depth $1/\bfD^s_{ij}$ is related to the vertex inverse depths $\bflambda$ through the barycentric coordinates \cite{Hughes2013CGP}, $\bfb_{ij}^\top \bflambda = 1/\bfD^s_{ij}$. Stacking these equations for all valid pixels $(i,j)$ in $\bfD^s$, we obtain:
\begin{equation} \label{eq:bary_depth}
\bfB \bflambda = \bfrho,
\end{equation}
where $\bfrho$ is a vector of the valid inverse depth measurements in $\bfD^s$ with elements $1/\bfD^s_{ij}$. Using Laplacian regularization as in \eqref{eq:loss_laplacian}, we formulate a least-squares problem in $\bflambda$:
%
%
\reviewerseven{
\begin{equation}
\bflambda^* = \arg\min_{\bflambda} \left( \|\bfB \bflambda - \bfrho\|_2^2 + w^{'}_{\bfV} \|\bfL_n \bflambda\|_2^2 \right).
\label{eq:depth_opt_linear}
\end{equation}}
The problem in \eqref{eq:depth_opt_linear} has a closed-form solution:
\begin{equation}
\bflambda^* = (\bfB^\top \bfB + w^{'}_{\bfV}\bfL_n^\top\bfL_n)^{-1}\bfB^\top \bfrho.
\label{eq:depth_opt_linear_solution}
\end{equation}
The regularization term, not only makes the initialized mesh smoother, but also guarantees that the solution exists even when the number of sparse depth measurements is smaller than the number of mesh vertices. Since the 2D mesh projection and $\bfL_n$ are pre-defined, the problem can be solved very efficiently, e.g., in less than $0.1$ sec for a mesh with $1024$ vertices. Given $\bflambda^*$, we obtain an initialized mesh $\mathcal{M}^{\text{int}}$ with each vertex coordinate as $[v^x_i/ \lambda_i^*, v^y_i/\lambda_i^*, 1/\lambda_i^*]$.
}

\subsection{Geometric Mesh Refinement}
\label{sec:geo_mesh_refine}

\begin{figure}[t]
  \centering
  \includegraphics[width=\linewidth,trim=0mm 8mm 0mm 5mm, clip]{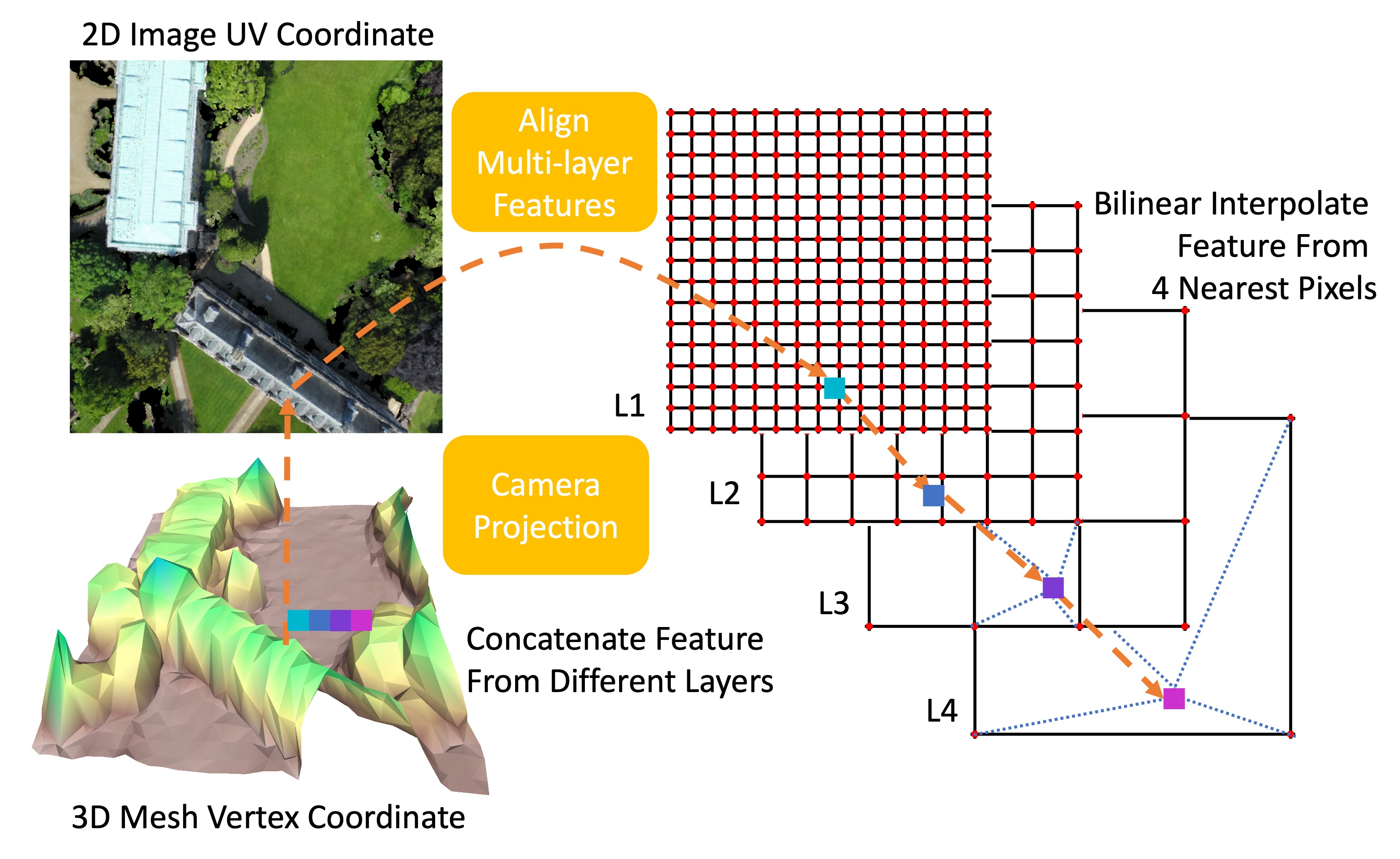}
  \caption{Illustration of image feature to mesh vertex association. With known camera intrinsics, each mesh vertex can be projected in uv coordinates (range $[0,1]$) onto the image plane. Bilinear interpolation is used to associate image feature maps at different resolutions with the mesh vertices. The features across different resolutions are concatenated to form a composite vertex feature.}
  \label{fig:vertex2image}
\end{figure}

Initialization using the sparse depth measurements only provides a reasonable mesh reconstruction but many details are missing. 
In the geometric refinement stage, we use a learning approach to extract features from both the 2D image and 3D initial mesh and regress mesh vertex spatial coordinate residuals. The ground-truth depth maps are used for supervision.

The photometric image information is useful for mesh refinement since man-made objects have sharp vertical surfaces, while natural terrain has noisy but limited depth variation. The sparse depth measurements also provide information about areas with large intensity variation. Inspired by Mesh R-CNN~\cite{Gkioxari2019MeshRCNN}, we design a network that extracts features from the 2D image, associates them with the 3D vertices of the initial mesh, and uses them to refine the vertex spatial coordinates. Our network has 3 stages: feature extraction, \reviewerb{vertex-image feature association}, and vertex graph convolution.



\noindent \textbf{Feature Extraction.}
\reviewerseven{We extract features from three sources: the RGB image $\bfI$, the rendered depth $\rho_D(\mathcal{M}^{\text{int}})$ from the initial mesh, and a Euclidean distance transform (EDT) $\bfE(\bfD^s)$ of the sparse depth measurements in the 2D image space, obtained by computing the Euclidean distance to the closest valid depth measurement pixel from each pixel coordinate. The three images are concatenated to form a 5-channel input (3-channels in $\bfI$, 1-channel in each $\rho_D(\mathcal{M}^{\text{int}})$ and $\bfE(\bfD^s)$):
\begin{equation}
\label{eq:2d_inputs}
    \bfF_{2D} = \text{concat}(\bfI, \rho_D(\mathcal{M}^{\text{int}}), \bfE(\bfD^s)).
\end{equation}
}
Four layers of features with different resolution and channels are extracted:%
\reviewerseven{
\begin{equation}
    [\bfL_1,\bfL_2,\bfL_3,\bfL_4] = \phi_{\text{res}}(\bfF_{2D};\bftheta_{2D}),
    \label{eq:feature_2D}
\end{equation}
where $\phi_{\text{res}}$ is a ResNet model~\cite{He2016resnet} with parameters $\bftheta_{2D}$.}

\noindent \reviewerb{\textbf{Vertex-Image Feature Association.}} 
Next, we construct 3D features for the mesh vertices by projecting each vertex to the image plane and interpolating the 2D image features. This idea is inspired by Pixel2Mesh \cite{Wang2018Pixel2Mesh}, which projects mesh vertices onto the image plane and extracts features at the projected coordinates. To obtain multi-scale features, we associate the projected mesh vertices with the intermediate layer feature maps $[\bfL_1,\bfL_2,\bfL_3,\bfL_4]$ from \eqref{eq:feature_2D}. The \reviewerb{vertex-image association} step is illustrated in Fig.~\ref{fig:vertex2image}. All features corresponding to different channels are concatenated to form composite vertex features. We define \reviewerb{$\text{associate}(\cdot,\cdot)$} as the function that assigns 2D features to 3D mesh vertices:
\begin{equation}
    \label{eq:vertex_associate}
    \bfV_{g_{\text{in}}} = \text{\reviewerb{associate}}(\mathcal{M},\phi_{\text{res}}(\bfF_{2D})),
\end{equation}
where $\bfV_{g_{\text{in}}} \in \mathbb{R}^{n\times(l_1+l_2+l_3+l_4)}$ are the vertex features and $l_i$ is the number of channels in feature map $\bfL_i$.



\noindent \textbf{Vertex Graph Convolution.} 
After the feature assignment, the mesh can be viewed as a graph with vertex features $\bfV_{g_{\text{in}}}$. Using the vertex features, a graph convolution network \cite{kipf2017semi,Gkioxari2019MeshRCNN} is a suitable architecture to predict coordinate deformation $\Delta \bfV$ for the vertex spatial coordinates to optimize the agreement between the refined mesh $\mathcal{M}^{\text{ref}} = (\bfV + \Delta \bfV)$ and the ground truth depth $\bfD$ according to the loss in \eqref{eq:final_loss}. To capture a larger region of feature influence, we use 3 layers of graph convolution $g^{\bfV}_1$,$g^{\bfV}_2$,$g^{\bfV}_3$, as follows:
\begin{equation}
\label{eq:vertex_offset}
\begin{aligned}
    \bfV^{\text{in}}_1 &= \text{ReLU}(\bfW^{\bfV}_1 \bfV_{g_{\text{in}}}) &\\
    \bfV^{\text{in}}_{i} &= \bfV^{\text{out}}_{i-1}, & i &= 2,3, \\
    \bfV^{\text{out}}_i &= \text{ReLU}(g^{\bfV}_i([\bfV^{\text{in}}_i;\bfV];\bftheta_{g\mathbf{V}i})), & i &= 1,2,3, \\
    \Delta \bfV &= \bfW^{\bfV}_2[\bfV^{\text{out}}_3;\bfV],
\end{aligned}
\end{equation}
where $\Delta \bfV \in \mathbb{R}^{n\times3}$ is the matrix of spatial coordinate residuals, $\bfW^{\bfV}_1,\bfW^{\bfV}_2$ are weight matrices of the linear layers, \reviewerseven{$\bftheta_{g\mathbf{V}i}$ are the graph convolution layer weights,} and ReLU is the Rectified Linear Unit activation function $\text{ReLU}(x) = \max(0,x)$.
\reviewerseven{The trainable parameters for vertex graph convolution are $\bftheta_{3D\bfV} = [\bfW^{\bfV}_1; \bfW^{\bfV}_2; \bftheta_{g\mathbf{V}1};\bftheta_{g\mathbf{V}2};\bftheta_{g\mathbf{V}3}]$.}
It is possible to concatenate more stages of \reviewerb{vertex-image feature association} and graph convolution. At stage $i$, the previous stage's refined mesh $\mathcal{M}^{\text{ref}}_{i-1}$ is set as the initial mesh $\mathcal{M}^{\text{int}}_i$ and new vertex features are extracted via \reviewerb{vertex-image feature association} and fed to new graph convolution layers. All refined meshes at different stages are evaluated using the ground-truth depth map $\bfD$ using the loss functions defined in \eqref{eq:final_loss}.

\subsection{Semantic Mesh Reconstruction}
\label{sec:semantic}

\begin{figure}[t]
  \centering
  \includegraphics[width=0.9\linewidth,trim=0mm 0mm 0mm 0mm, clip]{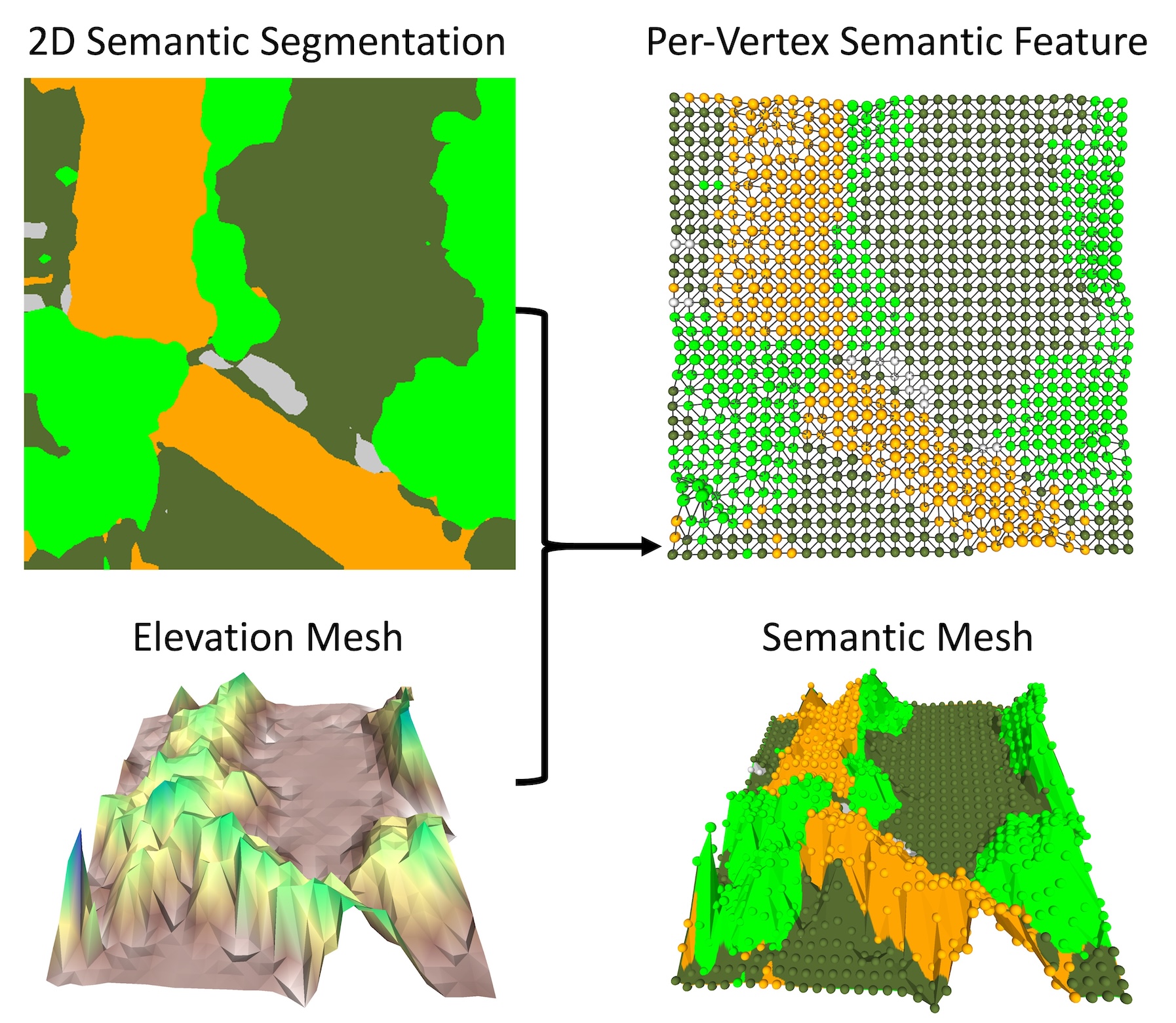}
  \caption{A 2D semantic segmentation feature map (top left) is used to generate semantic features for the 3D elevation mesh vertices (bottom left). Each mesh vertex is projected to the semantic segmentation feature map to retrieve an associated semantic feature (top right). Dense semantic features over the whole mesh can be obtained by interpolation on the mesh faces (bottom right).}
  \label{fig:mesh_sem_init}
\end{figure}

\begin{figure*}[t]
  \centering
  \includegraphics[width=0.9\linewidth,trim=0mm 0mm 0mm 0mm, clip]{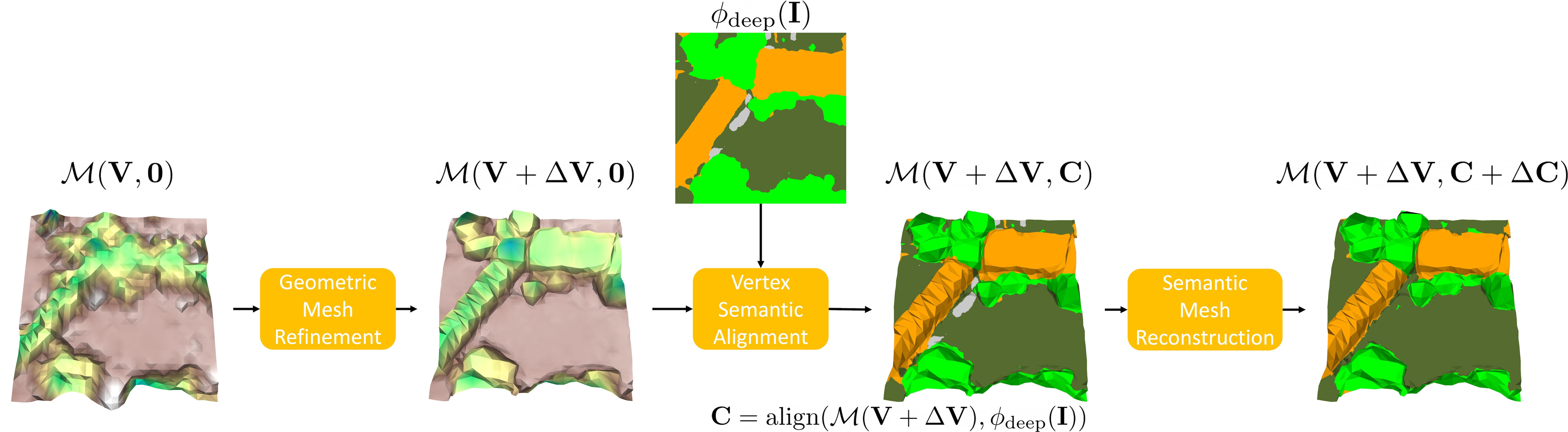}
  \caption{Mesh refinement stage: the mesh vertex spatial coordinate are refined to $\bfV + \Delta \bfV$ using graph convolution based on the RGB image features. Then, the semantic features $\bfC$ of the mesh vertices are initialized by projecting the vertices to the image plane and associating them with 2D semantic segmentation features. Finally, the vertex semantic features are refined to be $\bfC + \Delta \bfC$ using graph convolution.}
  \label{fig:refinement}
\end{figure*}

To further enrich the environment representation, we introduce semantic information in the mesh reconstruction. By assigning per-vertex semantic features $\bfC$, we can interpolate the semantic information over the whole mesh using barycentric coordinates (see Fig. \ref{fig:mesh_sem_init}). To obtain a 2D semantic segmentation image from the mesh, we use the differentiable semantic renderer $\rho_{\bfS}$ introduced in \eqref{eq:loss_sem_dice}. Both the vertex spatial coordinates $\bfV$ and the semantic features $\bfC$ can affect the rendered 2D semantic segmentation image $\rho_{\bfS}(\mathcal{M})$. Hence, by optimizing the semantic loss in \eqref{eq:loss_sem_dice}, we can refine both the semantic features and the geometric structure of the mesh.

\reviewerseven{We first obtain 2D semantic segmentation features $\phi_{\text{deep}}(\bfI;\bftheta_{2Dsem})$ from the RGB image $\bfI$ using the DeepLabv3 model \cite{Chen2017Deeplabv3} with parameters $\bftheta_{2Dsem}$.}
Then, we \reviewerb{associate} the mesh vertices to the 2D semantic feature map to get initial mesh vertex semantic features:
\begin{equation}
\label{eq:sem_vertex_align}
    \bfC = \text{\reviewerb{associate}}(\mathcal{M},\phi_{\text{deep}}(\bfI)).
\end{equation}
Fig.~\ref{fig:mesh_sem_init} illustrates the \reviewerb{mesh vertex association} with respect to the 2D semantic segmentation features. In the semantic refinement stage, we regress a semantic residual $\Delta \bfC$ for the semantic features. We use 3 layers of graph convolution $g^{\bfC}_1$,$g^{\bfC}_2$,$g^{\bfC}_3$. We also use the $\bfV_{g_{\text{in}}}$ extracted from \reviewer{ResNet} in \eqref{eq:vertex_associate} as an input to the first graph convolution layer. Additionally, we concatenate the initial mesh vertex semantic features $\bfC$ in \eqref{eq:sem_vertex_align} to the graph convolution input:
\begin{equation}
\label{eq:vertex_sem_offset}
\begin{aligned}
    \bfC^{\text{in}}_1 &= \text{ReLU}(\bfW^{\bfC}_1 \bfV_{g_{\text{in}}}) &\\
    \bfC^{\text{in}}_{i} &= \bfC^{\text{out}}_{i-1}, & i &= 2,3, \\
    \bfC^{\text{out}}_i &= \text{ReLU}(g^{\bfC}_i([\bfC^{\text{in}}_i;\bfV;\bfC];\bftheta_{g\mathbf{C}i})), & i &= 1,2,3, \\
    \Delta \bfC &= \bfW^{\bfC}_2[\bfC^{\text{out}}_3;\bfV;\bfC],
\end{aligned}
\end{equation}
where $\Delta \bfC \in \mathbb{R}^{n \times s}$ is the matrix of semantic residuals and $\bfW^{\bfC}_1,\bfW^{\bfC}_2$ are two matrices as linear layers. \reviewerseven{The trainable parameters for vertex semantic graph convolution are $\bftheta_{3D\bfC} = [\bfW^{\bfC}_1; \bfW^{\bfC}_2; \bftheta_{g\mathbf{C}1};\bftheta_{g\mathbf{C}2};\bftheta_{g\mathbf{C}3}]$.}

Now we can perform the joint geometric and semantic refinement. \reviewerseven{All trainable parameters for the 3D graph convolution are $\bftheta_{3D} = [\bftheta_{3D\bfV}; \bftheta_{3D\bfC}]$.} An illustration of the joint geometric and semantic refinement is provided in Fig. \ref{fig:refinement}. For the initial mesh $\calM(\bfV,\mathbf{0})$, we first estimate the geometric residuals $\Delta \bfV$ \eqref{eq:vertex_offset} from Sec.\ref{sec:geo_mesh_refine}. On the geometrically refined mesh $\calM(\bfV+\Delta \bfV,\mathbf{0})$, we initialize the semantic features as in \eqref{eq:sem_vertex_align} to get $\calM(\bfV+\Delta \bfV,\bfC)$. Then we estimate the semantic residuals $\Delta \bfC$ \eqref{eq:vertex_sem_offset}. The final joint geometric and semantic refined mesh is $\mathcal{M}^{\text{ref}} = (\bfV+ \Delta \bfV,\bfC + \Delta \bfC)$.

\subsection{\reviewera{Global Mesh Merging}}
\label{sec:mesh_merging}

\reviewerseven{
Given semantically annotated meshes obtained by our model from each camera view, a global mesh of the whole environment can be obtained by transforming each local mesh to the global frame using the camera pose trajectory and merging it into a combined global mesh. We design an approach to incrementally merge local meshes into a global mesh. Given a new local mesh obtained from a camera view with a known pose and the current global mesh, we update the global mesh by merging information from the local mesh.

First, we refine the global mesh vertices. We transform the global mesh to the local camera frame and project it onto the 2D image plane. If the resulting 2D global mesh covers an area over a certain threshold (e.g., 70\%), we regard the local frame as duplicate and proceed to the next frame. Otherwise, we determine the overlapping parts between the 2D projections of the global and local meshes. \reviewersix{We choose the vertices of the overlapped global mesh as a source point cloud and sample a point cloud from the overlapped local mesh vertices as a target point cloud. We perform non-rigid point cloud registration between the source and target point cloud using the Coherent Point Drift (CPD) algorithm \cite{Myronenko2010CPD}. Through this non-rigid transformation, we deform and refine the global mesh geometry based on the local mesh information.} 

Second, we introduce new vertices and faces into the global mesh from the non-overlapping region of the local mesh. We remove the faces of the local mesh that overlap with the global mesh projection on the image plane. Through this step, we decouple the global mesh and the local mesh because their 2D projections do not intersect with each other any longer. We perform 2D constrained Delaunay triangulation \cite{Anglada1997CDT} over the global and local mesh projections, keeping the edges of existing triangles in tact. Through this step, we connect the global mesh and the local mesh to obtain a new global mesh, which is lifted back to 3D using the vertex depth values and the known camera pose.

Fig.~\ref{fig:mesh_merge_2d} and Fig.~\ref{fig:mesh_merge} demonstrate the mesh merging process, which results in a single consistent global mesh and removes artifacts such as double layers in na\"ive mesh merging.

\begin{figure}[t]
  \centering
  \includegraphics[width=\linewidth,trim=0mm 0mm 0mm 0mm, clip]{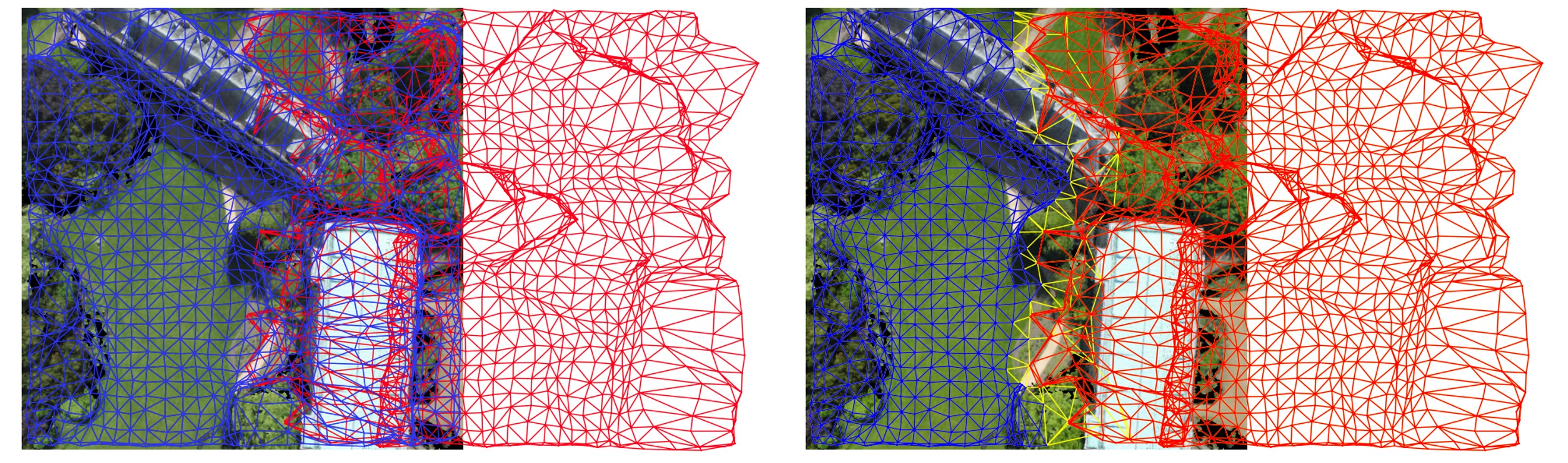}
  \caption{Red: global mesh. Blue: local mesh. Yellow: New edges after merging. Left: the refined global mesh overlaps with the local mesh. Right: the global and the local meshes are separated by removing overlapping faces and a new global mesh is generated via 2D constrained Delaunay triangulation.} 
  \label{fig:mesh_merge_2d}
\end{figure}    

\begin{figure}[t]
  \centering
  \includegraphics[width=\linewidth,trim=0mm 0mm 0mm 0mm, clip]{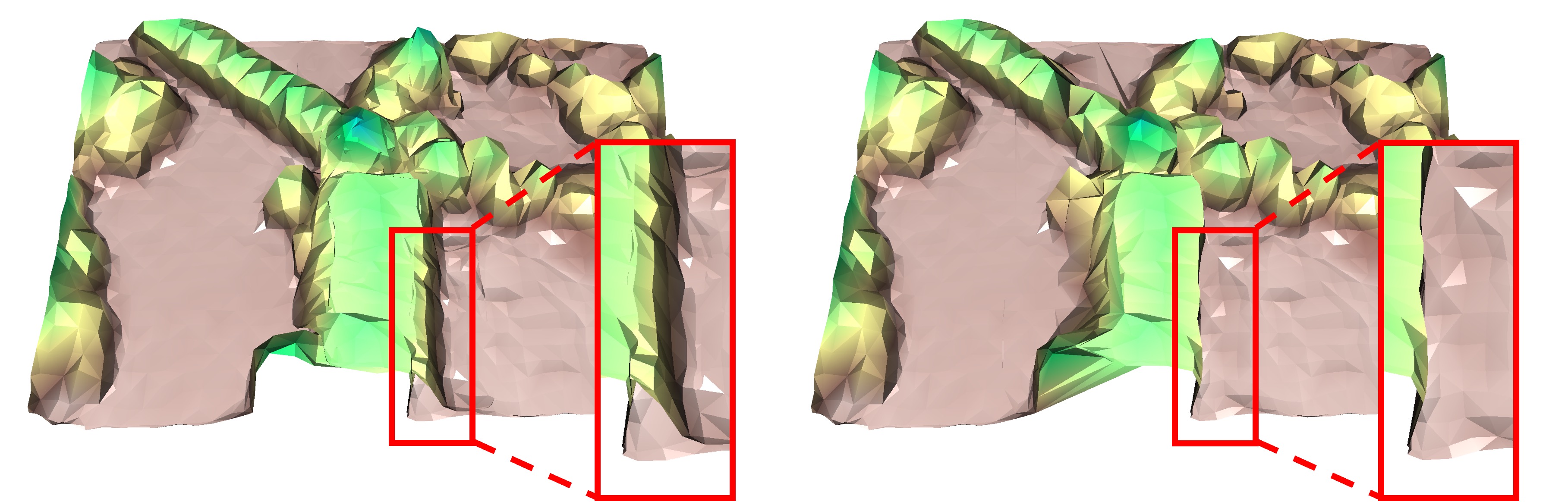}
  \caption{\reviewera{Left: Stacking two local meshes directly. Right: Merging two meshes with our proposed method in Sec.~\ref{sec:mesh_merging}.}} 
  \label{fig:mesh_merge}
\end{figure}
}

\section{Experiments}
\label{sec:experiments}

\begin{figure}[t]
  \centering
  \includegraphics[width=\linewidth,trim=0mm 0mm 0mm 0mm, clip]{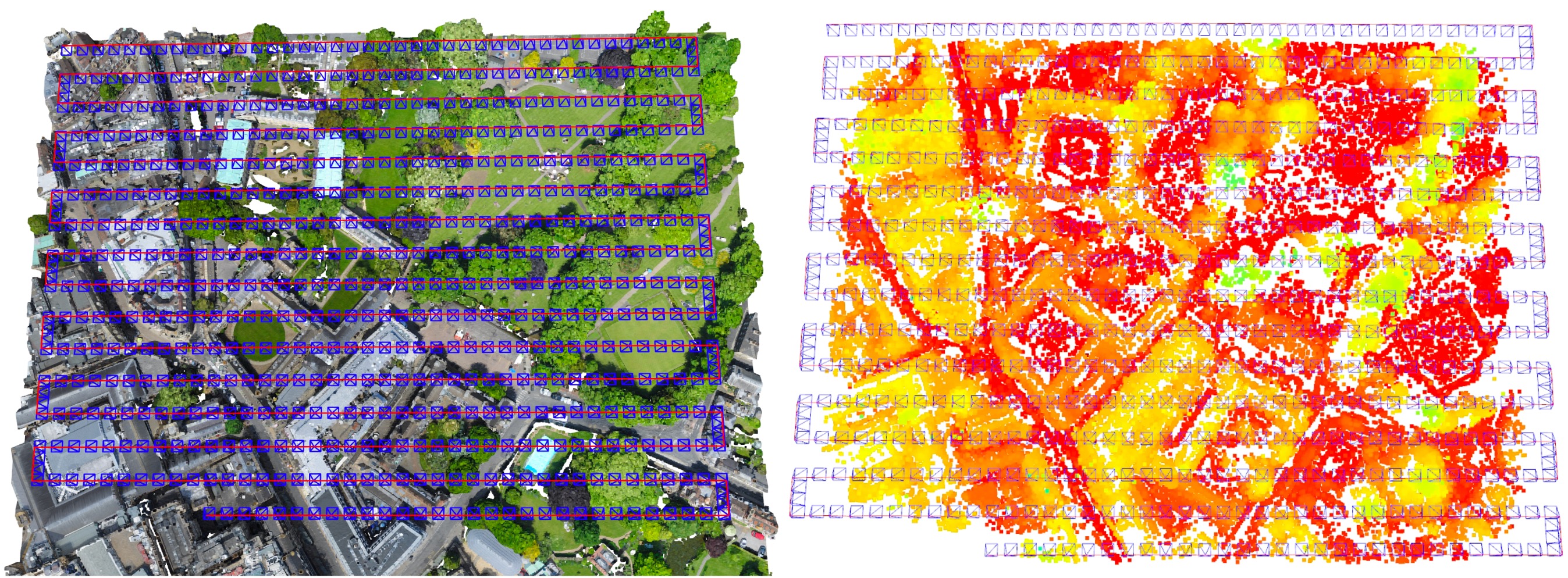}
  \caption{Left: Camera trajectory used to render RGBD images from a point cloud model generated from the SensatUrban dataset \cite{Hu2021Sensat}). Right: Sparse depth points and camera poses estimated by ORB-SLAM3 \cite{Campos2021ORB}. The color indicate elevation.}
  \label{fig:traj_pattern}
\end{figure}

\begin{figure*}[t]
  \centering
  \includegraphics[width=0.67\linewidth,trim=0mm 0mm 0mm 0mm, clip]{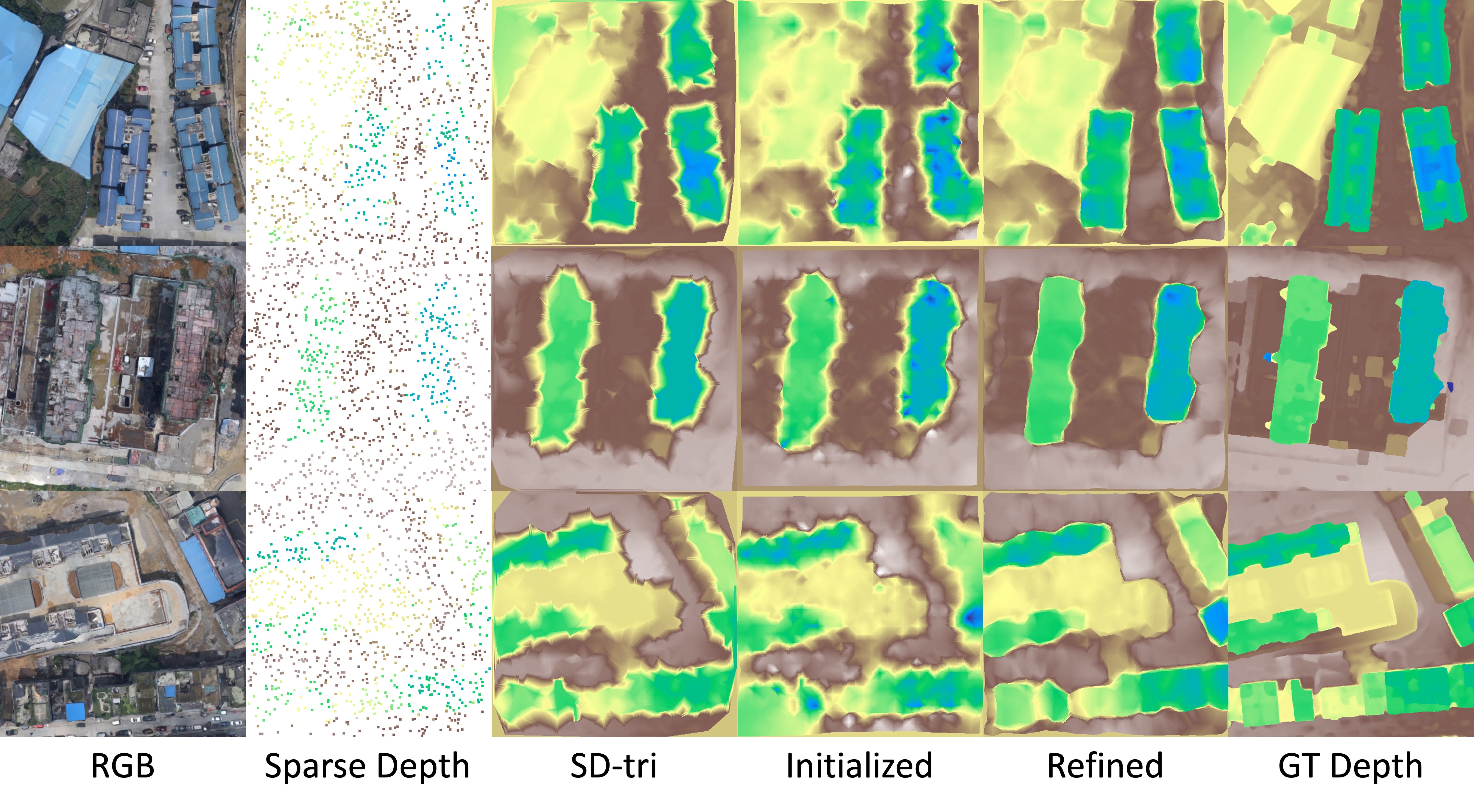}
  \caption{Mesh reconstructions on the WHU dataset~\cite{Liu2020WHU} visualized as rendered depth images. The colors indicate the relative depth values. Column 1: RGB images. Column 2: sparse depth measurements (around 1000). Column 3: meshes reconstructed from sparse-depth triangulation. Column 4: meshes after initialization (Sec.\ref{sec:mesh_init}). Column 5: meshes after neural network refinement (Sec.\ref{sec:geo_mesh_refine}). Column 6: ground-truth depth images.}
  \label{fig:whu_qual_results}
\end{figure*}

\begin{figure}[t]
  \centering
  \includegraphics[width=\linewidth,trim=0mm 10mm 0mm 10mm, clip]{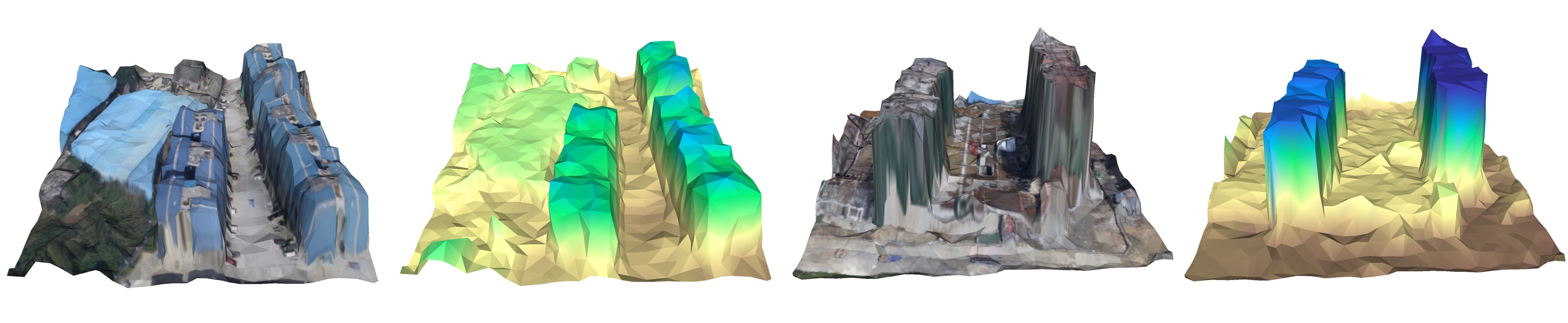}
  \caption{Reconstructed meshes painted with RGB texture and colors indicating elevations. These are associated with the first two rows in Fig. \ref{fig:whu_qual_results}. The sharp vertical transitions of the buildings are reconstructed accurately.}
  \label{fig:whu_qual_results_3d}
\end{figure}

\begin{figure}[t]
  \centering
  \includegraphics[width=\linewidth,trim=0mm 20mm 0mm 10mm, clip]{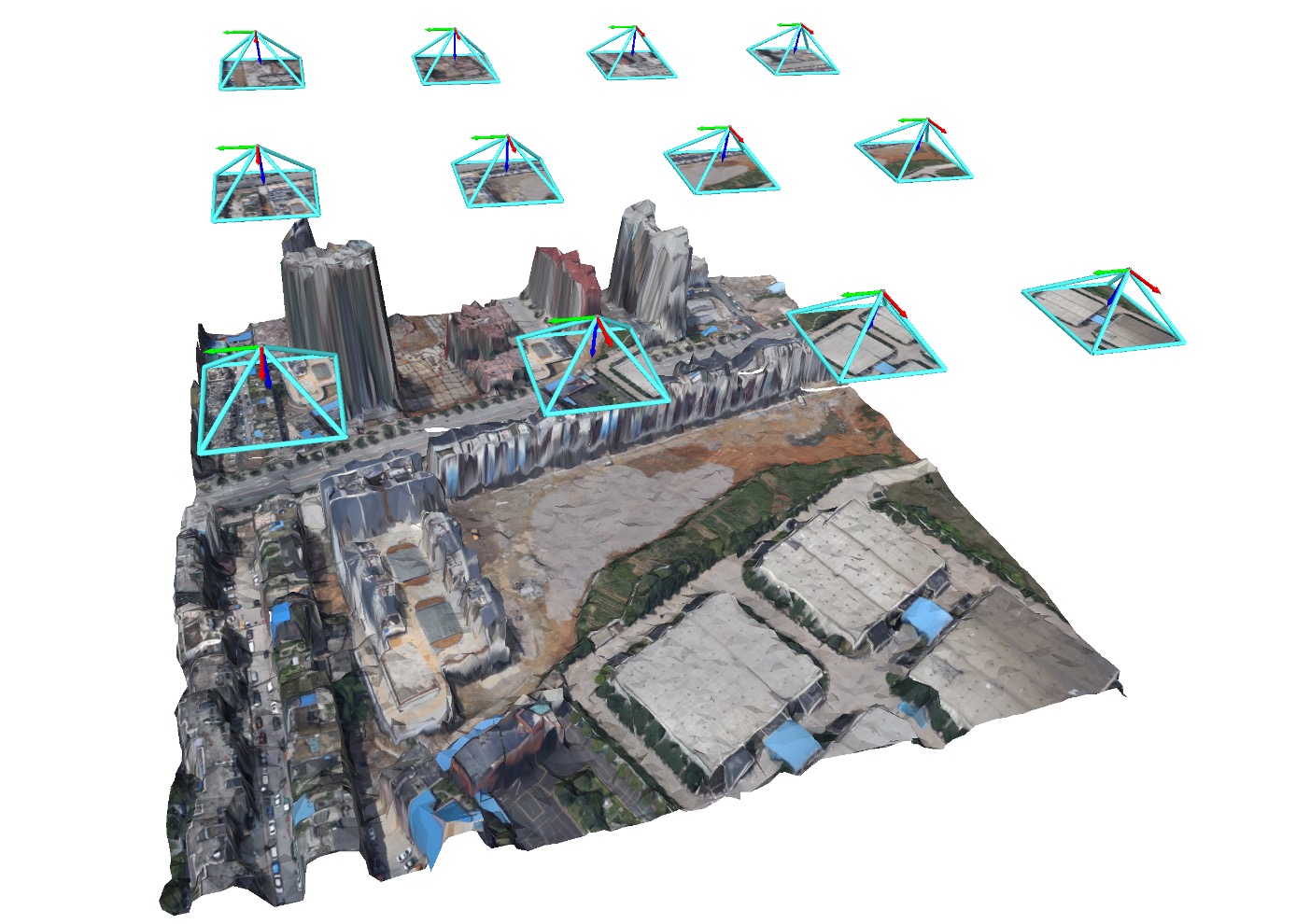}
  \caption{Complete environment model obtained by transforming to the global frame and merging local meshes from 12 camera views.}
  \label{fig:whu_global_map}
\end{figure}

\begin{figure*}[t]
  \centering
  \includegraphics[width=\linewidth,trim=0mm 0mm 0mm 0mm, clip]{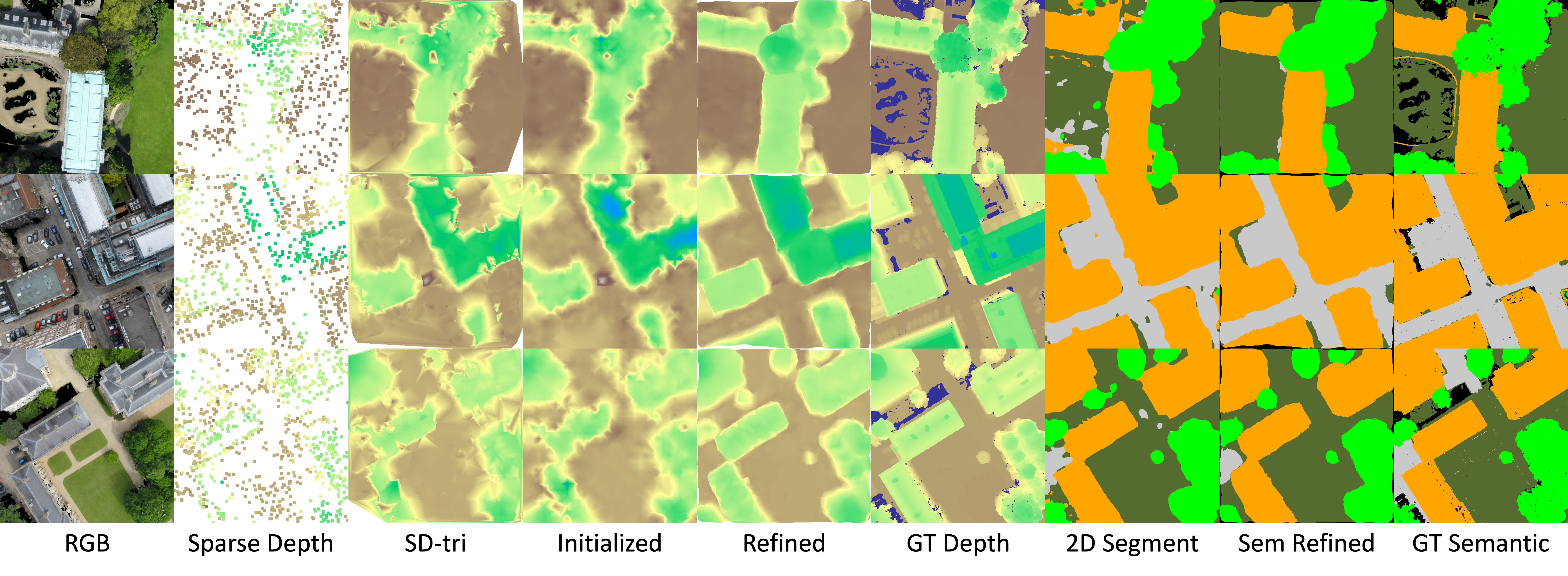}
  \caption{Mesh reconstructions on SensatUrban dataset \cite{Hu2021Sensat} visualized as rendered depth (colors indicate the relative depth values) and semantic images. The original 3D model is not fully complete so the RGB, GT Depth and GT Semantic may have little missing region. Column 1: RGB images. Column 2: sparse depth measurements (1000). Column 3: meshes reconstructed from sparse-depth triangulation. Column 4: meshes after initialization (Sec.\ref{sec:mesh_init}). Column 5: meshes after neural network refinement (Sec.\ref{sec:geo_mesh_refine}). Column 6: ground-truth depth images. Column 7: 2D semantic segmentation results. Column 8: meshes after neural network refinement (Sec.\ref{sec:semantic}). Column 9: ground-truth semantic segmentation maps.}
  \label{fig:sensat_qual_results}
\end{figure*}

\begin{figure}[t]
  \centering
  \includegraphics[width=\linewidth,trim=0mm 0mm 0mm 0mm, clip]{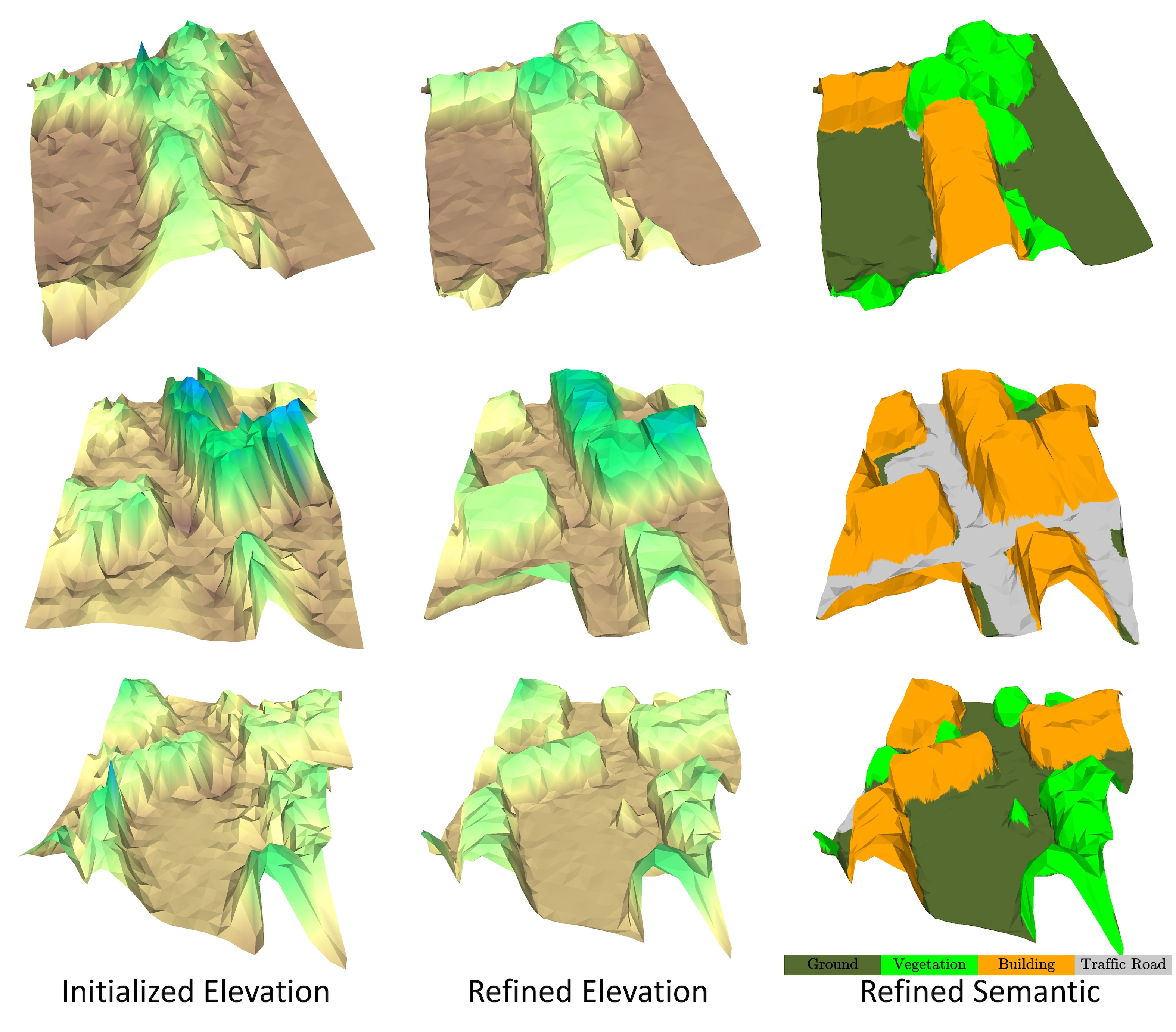}
  \caption{Reconstructed meshes painted with colors indicating elevations and semantic labels. These are associated with the three rows in Fig. \ref{fig:sensat_qual_results}. Column 1: initialized meshes. Column 2: refined meshes colored by elevation. Column 3: refined meshes colored by semantic categories.}
  \label{fig:sensat_qual_results_3d}
\end{figure}

\begin{figure*}[t]
  \centering
  \includegraphics[width=1\linewidth,trim=0mm 0mm 0mm 0mm, clip]{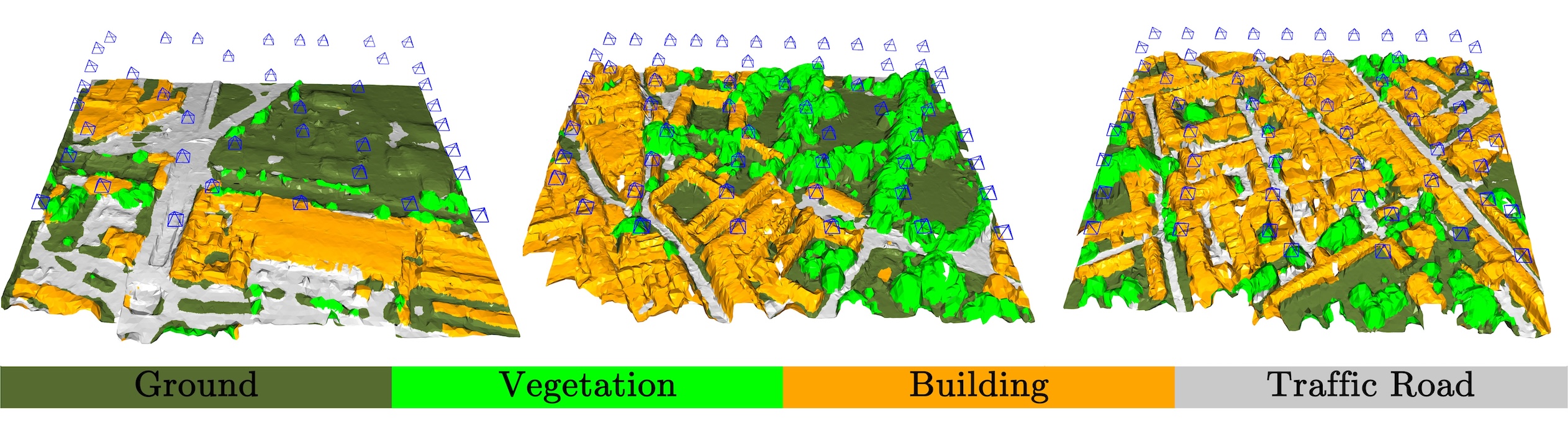}
  \caption{\reviewerthree{Global metric-semantic meshes reconstructed from three areas in the SensatUrban dataset \cite{Hu2021Sensat} by fusing the local mesh reconstructions at the keyframe camera poses (shown in blue). The three global meshes are obtained from 40/55/58 local keyframe meshes, respectively.}}
  \label{fig:sensat_qual_results_global}
\end{figure*}


\begin{table*}[t]
  \caption{Quantitative evaluation on the WHU dataset \cite{Liu2020WHU}. The second column shows the number of available sparse depth measurements per image and indicates whether the measurements are noisy (Sec.~\ref{sec:dataset}). The \emph{SD-tri} method triangulates a mesh using all the sparse depth measurements as vertices. The Regular-$n$ model generates a regular mesh with $n$ vertices and performs initialization and refinement steps as we propose in Sec. \ref{sec:technical_approach}. The \emph{Initialized} model constructs a mesh from the sparse depth (Sec.~\ref{sec:mesh_init}). The RGB, RGB+RD, RGB+RD+EDT methods refine the initialized mesh (Sec.~\ref{sec:geo_mesh_refine}), using different inputs respectively.}
  \label{tab:quantitative_WHU}
  \centering
  \begin{tabular}{|c|c|c|ccc|cccc|}
        \hline
        \multirow{2}{*}{Error} & Meshing & SD-tri & \multicolumn{3}{c|}{Regular-576} & \multicolumn{4}{c|}{Regular-1024} \\
        \cline{2-10}
        & Inputs & (vert = SD) & Initialized & RGB+RD & RGB+RD+EDT & Initialized & RGB & RGB+RD & RGB+RD+EDT \\
        \hline 
        \multirow{3}{*}{$\ell_2$} & 500 w/o noise & 1.492 & 2.069 & 1.670 & \textbf{1.637} & 1.861 & 1.575 & 1.289 & \textbf{1.252}\\
        & 1000 w/o noise & 1.172 & 1.834 & 1.596 & \textbf{1.546} & 1.535 & 1.298 & 1.124 & \textbf{1.097}\\
        & 2000 w/o noise & 0.916 & 1.941 & 1.551 & \textbf{1.511} & 1.344 & 1.144 & 1.045 & \textbf{1.024}\\
        \hline
        \multirow{3}{*}{$\ell_3$} & 500 w/o noise & 9.815 & 18.278 & \textbf{13.438} & 13.763 & 13.799 & 7.242 & \textbf{5.647} & 6.352\\
        & 1000 w/o noise & 6.494 & 17.762 & \textbf{12.938} & 13.574 & 11.872 & 5.876 & \textbf{4.911} & 5.703\\
        & 2000 w/o noise & 4.649 & 17.130 & \textbf{12.483} & 13.506 & 10.859 & 5.131 & \textbf{4.477} & 5.291\\
        \hline 
        \hline
        \multirow{3}{*}{$\ell_2$} & 500 & 1.865 & 2.294 & 1.809 & \textbf{1.768} & 2.155 & 1.828 & 1.486 & \textbf{1.456}\\
        & 1000 & 1.632 & 2.056 & 1.701 & \textbf{1.685} & 1.826 & 1.535 & 1.319 & \textbf{1.308}\\
        & 2000 & 1.485 & 1.717 & 1.655 & \textbf{1.654} & 1.629 & 1.364 & \textbf{1.236} & 1.241\\
        \hline
        \multirow{3}{*}{$\ell_3$} & 500 & 19.737 & 18.392 & \textbf{12.974} & 13.532 & 14.887 & 8.351 & \textbf{6.157} & 6.865\\
        & 1000 & 22.189 & 17.693 & \textbf{12.258} & 13.161 & 12.480 & 6.447 & \textbf{5.266} & 6.075\\
        & 2000 & 18.545 & 17.256 & \textbf{11.856} & 12.988 & 11.147 & 5.452 & \textbf{4.793} & 5.620\\
        \hline
  \end{tabular}
\end{table*}

\reviewerb{In this section, we evaluate our metric-semantic mesh reconstruction approach using aerial image sequences generated from three open-source 3D datasets: WHU MVS/Stereo \cite{Liu2020WHU}, SensatUrban \cite{Hu2021Sensat}, and STPLS3D \cite{Chen2022STPLS3D}. We evaluate the model generalization ability by training and testing on different datasets. Ablation studies are included to show the effectiveness of our choices in the model design.}



\subsection{Datasets}
\label{sec:dataset}

Our mesh reconstruction approach requires ground-truth depth and semantic segmentation data for supervised training, which are generally not available and challenging to obtain from RGB aerial images. We used photo-realistic point cloud models covering several km$^2$ reconstructed from real aerial images in WHU MVS/Stereo and SensatUrban dataset to render RGB, depth, and semantic segmentation images. This provides accurate depth and semantic supervision data, while keeping the RGB images realistic, which is important for real-world applications of our model.
\reviewerb{We also use data from the synthetic STPLS3D dataset, which has more variation in the scene layout and the texture.}
We divide the large point cloud models in each dataset into different regions
and generate different image sequences over them. Each camera trajectory follows a sweeping grid-pattern, which is common in drone flight planning (see Fig. \ref{fig:traj_pattern}). The camera trajectories are chosen to ensure enough image overlap for tracking and sparse depth reconstruction. RGBD images with resolution $512\times512$ are rendered along the trajectory from the ground-truth point cloud using PyTorch3D \cite{ravi2020pytorch3d}. We keep the RGB aerial image resolution at around $0.2$ meter/pixel. When semantic labels are available in the point cloud model, we also render semantic segmentation images with the same size as the RGBD images. All the data sequences are available in the Supplementary Material.


\noindent \textbf{WHU.} The WHU MVS/Stereo dataset~\cite{Liu2020WHU} provides geo-calibrated RGBD images rendered from a highly accurate 3D digital surface model of a $6.7 \times 2.2$ km$^2$ area over Meitan County, Guizhou Province, China. The 3D DSM model is not publicly available, so we recover a dense point cloud from the RGBD images as a ground-truth 3D model. Semantic labels are also not available in this dataset so we only perform geometric reconstruction using the WHU dataset. We obtain sparse depth measurements $\bfD^s$ for each image by applying OpenSfM~\cite{OpenSfM} to its four neighbor images with known camera intrinsic and extrinsic parameters. Since monocular structure from motion (SfM) suffers from scale ambiguity, we rescale the reconstructed point cloud obtained from OpenSfM to align it with the real 3D model. In reality the scale can be recovered from other sensor measurements like GPS or IMU. The point features reconstructed by OpenSfM are treated as sparse noisy depth measurements. The noise is due to feature detection and matching as well as the bundle adjustment step. We also obtain noiseless depth measurements with the same 2D sparsity pattern from the ground-truth depth images $\bfD$. We vary the number of available sparse depth measurements as 500, 1000, 2000. We generate 20 camera trajectory sequences with 200 images in each sequence, split into 14 for training, 2 for validation, and 4 for testing.

\noindent \textbf{SensatUrban.} The SensatUrban dataset \cite{Hu2021Sensat} is a point cloud dataset obtained using photogrammetry in two urban areas in Birmingham and Cambridge, UK. Each 3D point in the dataset is labeled as one of 13 semantic classes. The Birmingham region covers an area of $1.2$ km$^2$. The Cambridge region covers an area of $3.2$ km$^2$. We only use the training set part of the data in which point cloud semantic labels are available. We keep 4 semantic categories (ground, vegetation, building, and traffic road) and merge or discard the remaining less-frequent categories. We used monocular ORB-SLAM3 \cite{Campos2021ORB} to estimate the camera poses and sparse feature depths on the SensatUrban dataset. Compared with OpenSfM, ORB-SLAM3 performs sequential optimization of the image sequences, instead of looping over all the images to find matching pairs. As a result, it runs faster (1-10 Hz) and may be deployed on an aerial robot directly. 
\reviewer{We use OrbSLAM3 in order to ensure that our method can operate incrementally in time and handle pose and sparse depth estimation errors typical for online SLAM algorithms.}
We also re-scale the reconstructed point cloud and camera poses to align with the real 3D model. Finally, we project the point cloud to each camera frame to derive a sparse depth image. We vary the number of available sparse depth measurements as 500, 1000, 2000, 4000. We generate 13 camera trajectory sequences with 660 images in each sequence, split into 8 for training, 2 for validation, and 3 for testing. 

\reviewerb{\noindent \textbf{STPLS3D.} The STPLS3D dataset \cite{Chen2022STPLS3D} is a richly-annotated synthetic 3D aerial photogrammetry point cloud dataset with more than 16 km$^2$ of landscapes and up to 18 fine-grained semantic category annotations. To ensure the object placements in the virtual environments resemble real city blocks, the environments are built based on Geographic Information System (GIS) data that are publicly available. We use the same 4 semantic categories as for the SensatUrban dataset. We generated 38 camera trajectory sequences with 660 images in each sequence, split into 26 for training, 4 for validation, and 8 for testing. Camera keyframe poses and sparse depth measurements were estimated using ORB-SLAM3.}

\subsection{Implementation Details}
\label{sec:impl_details}

During training, we use 1000 sparse depth measurements per image and generate a mesh model with 576/1024/2025 vertices. 
For the WHU dataset, the weights of the loss function in \eqref{eq:final_loss} are set to $[w_2, w_3, w_{\bfV}, w_{\calE},w_{\bfS},w_{\bfC}] = [3,1,0.5,0.01,0,0]$. For \reviewer{the SensatUrban and the STPLS3D dataset}, the weights are set to $[w_2, w_3, w_{\bfV}, w_{\calE},w_{\bfS},w_{\bfC}] = [5,1,0.5,0.01,5,0.5]$ for the joint geometric-semantic training (Sec. \ref{sec:semantic}). For the geometric training (Sec. \ref{sec:geo_mesh_refine}), the last two weights are set to be $0$. 
\reviewerc{\reviewerseven{The loss weights are decided through evaluation on the validation set. We use the 2D loss $\ell_{2}$ in \eqref{eq:loss_2D} as the metric to choose the best model on the validation set.}}
The Chamfer distance $d$ in the $\ell_3$ loss \eqref{eq:loss_3D} is computed using 10000 samples. 

For the WHU experiments in Sec.~\ref{sec:exp_WHU}, we use ResNet-18 for the 2D feature extraction. For the remaining ones including the generalization experiments, we use ResNet-34. The ResNet is initialized with the pretrained weights on ImageNet-1K. The ResNet and GCN parameters of our model are optimized jointly during the mesh refinement training using the Adam optimizer \cite{kingma2014adam} with initial learning rate of $0.0005$ for $100$ epochs. For the semantic reconstruction task, we first train a DeepLabv3 model with ResNet-50 backbone \cite{Chen2017Deeplabv3} alone for 2D semantic segmentation on the SensatUrban training set. The ResNet-50 is initialized with the pretrained weights on ImageNet-1K. We use the Cross Entropy loss for training and set the class weight as [ground, vegetation, building, traffic road] = [$1,2,3,3$]. During the mesh semantic refinement step, we use the Dice loss in \eqref{eq:loss_sem_dice} and keep the same per-class weights. We use three graph convolution stages for the WHU dataset and two graph convolution stages for \reviewer{the SensatUrban and the STPLS3D dataset}.
For the joint geometric-semantic training, we concatenate two graph convolution stages, where the first stage predicts the geometric residual only and the second stage predicts both the geometric and the semantic residuals. 
\reviewerseven{All trainable parameters of the model in \eqref{eq:problem} are $\bftheta = [\bftheta_{2D};\bftheta_{2Dsem};\bftheta_{3D}]$, including the parameters of the 2D features extraction, 2D semantic segmentation, and 3D graph convolution models.}

\subsection{Geometric Reconstruction}
\label{sec:exp_WHU}

Our experiments report the $\ell_2$ error in \eqref{eq:loss_2D} and the $\ell_3$ error in \eqref{eq:loss_3D} for the reconstructed meshes. The $\ell_2$ error emphasizes the accuracy of the projected depth, while $\ell_3$ emphasizes the regions of large depth variation. 

For comparison, we define a baseline method that triangulates the sparse depth measurements directly to build a mesh. The baseline method performs Delaunay triangulation on the 2D image plane over the depth measurements and projects the flat mesh to 3D using the measured vertex depths. We refer to the baseline method as sparse-depth-triangulation (\emph{SD-tri}). SD-tri defines vertices at all sparse depth measurements (500, 1000, or 2000) and, hence, may produce meshes with different number of vertices compared to other models.

\reviewerb{First, we perform geometric reconstruction on the WHU dataset.} The quantitative results from the comparison are reported in Table \ref{tab:quantitative_WHU}. All models are trained with 1000 sparse depth measurements and directly generalize to different numbers of sparse depth measurements. We compared three options for the 2D inputs provided to the mesh refinement stage: an RGB image only (RGB, 3-channels), an RGB image plus rendered depth from the initial mesh (RGB+RD, 4-channels), and an RGB image plus rendered depth from the initial mesh plus Euclidean distance transform (EDT) obtained from of the sparse depth measurements (RGB+RD+EDT, 5-channels). The model using RGB-only does not perform as well as the other two. The RGB+RD+EDT model has the best performance according to the $\ell_2$ error metric. The RGB+RD method has similar performance in the $\ell_2$ metric and smaller $\ell_3$ error compared to RGB+RD+EDT. The RGB+RD model is used to generate our qualitative results in Fig.~\ref{fig:whu_qual_results}, \ref{fig:whu_qual_results_3d}, \ref{fig:whu_global_map} with 1024-vertex meshes because it offers good performance according to both error metrics.



\reviewera{At the bottom of Table~\ref{tab:quantitative_WHU}, we evaluate the mesh reconstruction accuracy with noisy sparse depth measurements obtained from OpenSfM. The measurements are noisy due to feature matching errors, local minima during bundle adjustment, and the simple projective camera model used for optimization. The average per image errors of the 500/1000/2000 sparse depth measurements were 1.011/1.017/1.023 meters, respectively.} The baseline SD-tri method performs well in a noiseless setting but degenerates drastically when noise from the SfM feature reconstruction is introduced. In contrast, our model is more robust to noise due to two factors. First, our mesh initialization and refinement stages both include explicit mesh regularization terms (in \eqref{eq:loss_laplacian} and \eqref{eq:loss_edge}). Second, the image features extracted during the mesh refinement process help distinguish among different terrains and structures. The latter is clear from the improved accuracy of the refined, compared to the initialized, meshes. We also report the performance using a mesh with only 576 vertices. When the depth measurements are noisy, the 576-vertex mesh has lower $\ell_2$ loss compared with the baseline method with similar number of vertices. It even has lower $\ell_3$ loss compared with meshes with more vertices generated from the baseline method.

Qualitative results are presented in Fig.~\ref{fig:whu_qual_results} and \ref{fig:whu_qual_results_3d}. Compared with SD-tri and initialized meshes, the refined meshes have smoother boundaries on the side surfaces of the buildings. The guidance from the image features allows the refined meshes to fit the 3D structure better. Fig.~\ref{fig:whu_global_map} shows a global mesh reconstruction obtained by transforming and merging 12 camera-view mesh reconstructions. The local meshes are transformed to global frame using the camera keyframe poses and no post-processing is used to merge them into a single global mesh.

\begin{table*}[t]
  \caption{\reviewer{Quantitative evaluation on the SensatUrban dataset. The second column shows the number of available sparse depth measurements per image (Sec.~\ref{sec:dataset}). The baseline \emph{SD-tri} method triangulates a mesh using all sparse depth measurements as vertices. The Regular-$n$ model generates a regular mesh with $n$ vertices and performs initialization and refinement steps (Sec. \ref{sec:technical_approach}).} 
  }
  \label{tab:quantitative_Sensat}
  \centering
  \begin{tabular}{|c|c|c|cc|cc|cc|}
        \hline
        \multirow{2}{*}{Error} & Meshing & SD-tri & \multicolumn{2}{c|}{Regular-576} & \multicolumn{2}{c|}{Regular-1024} & \multicolumn{2}{c|}{Regular-2025}\\
        \cline{2-9}
        & Inputs & (vert = SD) & Initialized & Refined & Initialized & Refined & Initialized & Refined \\
        \hline
        \multirow{4}{*}{$\ell_2$} & 500 & 2.018 & 2.050 & 1.175 & 2.204 & \textbf{1.088} & 1.992 & 1.171\\
        & 1000 & 1.843 & 1.841 & 1.096 & 1.865 & \textbf{1.000} & 2.033 & 1.153\\
        & 2000 & 1.715 & 1.796 & 1.120 & 1.700 & \textbf{0.988} & 1.752 & 1.209\\
        & 4000 & 1.647 & 1.834 & 1.181 & 1.662 & \textbf{1.026} & 1.630 & 1.283\\
        \hline
        \multirow{4}{*}{$\ell_3$} & 500 & 8.926 & 7.898 & 2.371 & 9.871 & \textbf{2.128} & 7.645 & 2.371\\
        & 1000 & 7.796 & 6.353 & 2.075 & 6.725 & \textbf{1.815} & 8.875 & 2.284\\
        & 2000 & 7.164 & 5.989 & 2.133 & 5.527 & \textbf{1.745} & 6.200 & 2.500\\
        & 4000 & 6.908 & 6.176 & 2.339 & 5.217 & \textbf{1.844} & 5.364 & 2.806\\
        \hline
  \end{tabular}
\end{table*}

\begin{table}[t]
  \caption{\reviewer{Semantic segmentation per-class IoU for different geometric-semantic models. The definitions of the different models can be found in Sec. \ref{sec:exp_SensatUrban} and Sec. \ref{sec:ablation}.}}
  \label{tab:sem_sensat}
  \centering
  \begin{tabular}{|c|cccc|}
        \hline
        Class & Ground & Vegetation & Building & Traffic Road\\
        \hline
        Geo Init & 0.642 & 0.810 & 0.846 & 0.643 \\
        
        Geo Refine & 0.644 & 0.809 & 0.840 & 0.644 \\
        \hline 
        Cross Entropy & 0.661 & 0.805 & 0.843 & 0.660 \\
        
        Focal & 0.663 & 0.807 & 0.844 & 0.657 \\
        
        Jaccard & 0.664 & \textbf{0.826} & \textbf{0.863} & 0.653 \\
        
        \hline
        Our Model (Dice) & \textbf{0.674} & 0.824 & 0.860 & \textbf{0.663} \\
        \hline
        2D Seg & 0.649 & 0.834 & 0.854 & 0.648 \\
        \hline
  \end{tabular}
\end{table}

\begin{table}[t]
  \caption{\reviewer{Geometric error for different metric-semantic models. The definitions of the different models and loss functions can be found in Sec.~\ref{sec:exp_SensatUrban} and Sec.~\ref{sec:ablation}.}}
  \label{tab:geo_sensat}
  \centering
  \begin{tabular}{|c|cc|}
        \hline
        Method & Depth ($\ell_2$) & Chamfer ($\ell_3$)\\
        \hline
        Geo Init & 1.866 & 6.725 \\
        Geo Refine & 1.000 & 1.815\\
        \hline
        Cross Entropy & 0.994 & 1.793 \\
        Focal & 1.035 & 1.912 \\
        Jaccard & \textbf{0.976} & 1.776 \\
        \hline
        Our Model (Dice) & \textbf{0.976} & \textbf{1.763} \\
        \hline
  \end{tabular}
\end{table}

\subsection{Joint Geometric \& Semantic Reconstruction}
\label{sec:exp_SensatUrban}

On the SemsatUrban dataset, we first perform geometric reconstruction with the same settings as in the WHU dataset. We train three models with different numbers of mesh vertices: $576 = 24^2$, $1024=32^2$ and $2025=45^2$. The quantitative results are reported in Table \ref{tab:quantitative_Sensat}. As the number of sparse depth measurements increases, the baseline SD-tri method has better accuracy because the number of mesh vertices also increases. Our initialized meshes with fewer vertices are comparable with the SD-tri mesh, and the refined meshes are much better, especially according to the 3D metric $\ell_3$. This shows that the joint 2D-3D loss in \eqref{eq:final_loss} enables our model to capture 3D structure details. Comparing the number of input depth measurements, we find that around 2000 measurements on the $512\times512$ image provide the best performance, while more do not noticeably improve the results. Regarding the number of mesh vertices, all three mesh sizes perform well. While we can see that the 1024-vertex mesh is generally better than 576-vertex mesh, the 2025-vertex mesh does not show an advantage over the 1024-vertex mesh. This indicates that good accuracy can be achieved with a light-weight storage-efficient mesh model. 

We choose the 1024-vertex mesh to perform joint geometric-semantic reconstruction using 1000 sparse depth measurements. To evaluate the semantic reconstruction, we render a 2D semantic image from the mesh reconstruction and calculate the per-class Intersection over Union (IoU). For comparison, we report the IoU of the DeepLabv3 2D semantic segmentation model (named 2D Seg), the direct projection of the 2D semantic segmentation image onto the initial mesh as in \eqref{eq:sem_vertex_align} (named Geo Init) and the semantic segmentation projection onto the geometrically-refined mesh (named Geo Refine). Only 2D Seg is using a dense semantic image while the other methods store semantic features on the mesh vertices and interpolate through the semantic mesh renderer. As we can see in Table \ref{tab:sem_sensat}, our semantic residual refinement model improves the semantic segmentation performance compared to the direct projection of the 2D semantic segmentation image. Our approach also outperforms 2D Seg on most of the categories (ground, building, traffic Road) even thought it is using only 0.4\% of the points to store the semantic information (1024 mesh vertices vs $512 \times 512$ segmentation image). 
\reviewerthree{Using a compact mesh with few vertices improves the computation and memory efficiency but restricts the mesh from modeling small regions (e.g., small objects like cars, street furniture). Our approach can capture additional semantic categories if the number of mesh vertices is increased, e.g., to $64^2=4096$ or more, and a semantic segmentation network capable of segmenting small regions is used. Achieving this does not require any changes to the model architecture but only retraining the model parameters.}

Further, we investigate whether the semantic mesh refinement affects the geometric reconstruction quality. In Table \ref{tab:geo_sensat}, we can see that our joint geometric-semantic mesh reconstruction achieves better geometric accuracy compared with purely geometric training. This can be explained by the fact that the semantic category information serves as regularization for the geometric properties. The results show that the geometric and semantic information help each other. More qualitative results for single-image reconstruction are provided in Fig. \ref{fig:sensat_qual_results} and \ref{fig:sensat_qual_results_3d}. Compared with SD-tri and initialized mesh, the refined mesh achieves higher reconstruction accuracy. The semantic refinement can improve the 2D semantic segmentation results. We can see that some noisy classification labels are removed after the refinement. 

\reviewerthree{To evaluate our global mesh-merging method based on CPD and Delaunay triangulation, we compare it with a simple stacking method that transforms the same set of local meshes to the global frame and simply treats them as a single global mesh. For each scene, about 10\% of the frames are used to generate a global mesh. The global mesh is rendered with respect to all frames, including ones that were not used for its construction, to compute the error. Therefore, the error is larger than the per-frame prediction evaluation. The quantitative results are reported in Table~\ref{tab:mesh_merge}. The stacked global mesh has many double layers in regions where the local meshes overlap while our method successfully merges the local meshes into a consistent global mesh, which improves the reconstruction accuracy, especially according to the 3D Chamfer error metric. Fig.~\ref{fig:sensat_qual_results_global}, shows the reconstruction of three global metric-semantic meshes using data from the SensatUrban dataset \cite{Hu2021Sensat}.}

    \begin{table}[t]
      \caption{\reviewerthree{Geometric error for global mesh-merging methods.}}
      \label{tab:mesh_merge}
      \centering
      \begin{tabular}{|c|cc|}
            \hline
            Method & Depth ($\ell_2$) & Chamfer ($\ell_3$)\\
            \hline
            Simple Stacking & 1.620 & 6.267 \\
            Our Method & 1.600 & 5.219 \\
            \hline
      \end{tabular}
    \end{table}

\reviewerb{We also evaluated our model on the synthetic STPLS3D dataset. We use a 1024-vertex mesh to perform geometric-only and joint geometric-semantic reconstruction using 1000 sparse depth measurements per image. The quantitative results are shown in Table \ref{tab:geo_stpls3d}. Our joint geometric-semantic mesh reconstruction method out-performs geometric-only mesh reconstruction, which verifies the effectiveness of fusing both geometric and semantic information.}

\begin{table}[t]
  \caption{\reviewerb{Geometric error on the STPLS3D dataset.}}
  \label{tab:geo_stpls3d}
  \centering
  \begin{tabular}{|c|cc|}
        \hline
        Method & Depth ($\ell_2$) & Chamfer ($\ell_3$)\\
        \hline
        SD-tri & 2.039 & 11.832 \\
        Geo Init & 1.996 & 10.417 \\
        Geo Refine & 0.977 & 2.807 \\
        Sem Refine & 0.972 & 2.725 \\
        \hline
  \end{tabular}
\end{table}

\reviewerb{
\subsection{Generalization Across Datasets}
\label{sec:generalization}
In this section, we evaluate the generalization ability of our model, trained on one dataset and applied to another. To align the datasets, we regenerated the RGB images and the sparse depth measurements on the WHU dataset to follow the same camera intrinsic parameters and trajectory patterns in the SensatUrban and the STPLS3D dataset so that there are 660 frames for each trajectory and ORB-SLAM3 is used to estimate sparse depth measurements and camera keyframe poses. It is challenging to achieve zero-shot generalization, so we also include a finetuning step. During finetuning, we only use 10\% of the target domain training set and train for 30 epochs. A validation set (10\% of the target domain validation set) is used to choose the best model with the 2D loss $\ell_2$ as the metric. Usually, models trained on larger datasets show better generalization ability. We choose to use a model trained on WHU to generalize to SensatUrban and a model trained on STPLS3D to generalize to both WHU and SensatUrban. The average per image sparse depth errors in meters for 1000 depth samples were 1.771 on STPLS3D, 2.460 on WHU, and 1.597 on SensatUrban. The sparse depth measurements are generated through ORB-SLAM3.


First, we evaluate how the model trained on STPLS3D generalizes to WHU. The geometric error is reported in Table \ref{tab:geo_generalization_whu}. The WHU dataset is more challenging due to the presence of denser and taller ($>30$m) buildings. The camera intrinsics and the flight pattern and height are different compared to the data generated in Sec. \ref{sec:exp_WHU} so the numbers in Table \ref{tab:geo_generalization_whu} are not directly comparable with Table \ref{tab:quantitative_WHU}. 
Zero-shot generalization does not work for WHU, which is understandable given the large domain gap. The STPLS3D synthetic dataset uses scene layouts extracted from a U.S. Geological Survey (USGS) which covers cities in the United States, while the WHU dataset is collected in a Chinese city. After finetuning with only 10\% of the original WHU training set, our model generalizes well to WHU, and even outperforms the model trained purely on WHU.

Next, we evaluate how models trained on WHU and STPLS3D generalize to SensatUrban. The results are presented in Table \ref{tab:geo_generalization_sensaturban} and Table \ref{tab:sem_generalization_sensaturban}. We report only geometric error for the model trained on WHU. We can see that zero-shot generalization from WHU to SensatUrban improves the initialized meshes, while a finetuned model performs even better. We train a geometric-only model and a metric-semantic model on STPLS3D. In terms of geometric loss, both STPLS3D models generalize well to SensatUrban and their performance after finetuning is close to that of a model trained on SensatUrban. However, the metric-semantic model is slightly worse than the pure geometric model. In terms of semantic segmentation performance, zero-shot generalization does not perform well and especially fails on the traffic road category. After finetuning, the metric-semantic model can largely close the gap between itself and the SensatUrban model. Given that the RGB images from the synthetic scenes in STPLS3D have very different appearance, it is understandable that the semantic model that heavily relies on the RGB image might be harder to generalize compared to the geometric-only model.

These experiments demonstrate promising generalization ability of our mesh reconstruction method, using limited data to finetune or even without finetuning in some cases. The model generalizes better in terms of geometric reconstruction than in terms of semantic classification. Nevertheless, it is exciting to see that a model trained on a synthetic dataset (STPLS3D) can generalize well to real data. This makes it possible to achieve good performance by training a model with inexpensive synthetic data that comes with free ground-truth labels and finetuning on a small set from the target domain.}


\begin{table}[t]
  \caption{\reviewerb{Generalization experiment: Geometric error on WHU. Brackets indicate the dataset trained on.}}
      \label{tab:geo_generalization_whu}
  \centering
  \begin{tabular}{|c|cc|}
        \hline
        Method & Depth ($\ell_2$) & Chamfer ($\ell_3$)\\
        \hline
        SD-tri & 3.628 & 40.431 \\
        Init & 3.582 & 38.570 \\
        Refine (WHU) & 2.253 & 11.332 \\
        \hline
        Refine (STPLS3D) & 20.043 & 1478.478 \\
        Refine (STPLS3D finetune) & 2.047 & 10.796 \\
        \hline
  \end{tabular}
\end{table}

\begin{table}[t]
  \caption{\reviewerb{Generalization experiment: Geometric error on SensatUrban. Brackets indicate the dataset trained on.}}
  \label{tab:geo_generalization_sensaturban}
  \centering
  \begin{tabular}{|c|cc|}
        \hline
        Method & Depth ($\ell_2$) & Chamfer ($\ell_3$)\\
        \hline
        SD-tri & 1.843 & 7.796 \\
        Init & 1.865 & 6.725 \\
        Refine (SensatUrban) & 1.000 & 1.815 \\
        Sem Refine (SensatUrban) & 0.976 & 1.763 \\
        \hline
        Refine (WHU) & 1.439 & 3.827 \\
        Refine (WHU finetune) & 1.112 & 2.223 \\
        \hline
        Refine (STPLS3D) & 1.442 & 3.973 \\
        Sem Refine (STPLS3D) & 1.501 & 4.309 \\
        Refine (STPLS3D finetune) & 1.021 & 1.992 \\
        Sem Refine (STPLS3D finetune) & 1.043 & 2.111 \\
        \hline
  \end{tabular}
\end{table}

\begin{table}[t]
  \caption{\reviewerb{Generalization experiment: Semantic segmentation per-class IoU on SensatUrban. Brackets indicate the dataset trained on.}}
  \label{tab:sem_generalization_sensaturban}
  \centering
  \resizebox{\linewidth}{!}{\begin{tabular}{|c|cccc|}
        \hline
        Class & Ground & Vegetation & Building & Traffic Road\\
        \hline
        2D Seg (SensatUrban) & 0.649 & 0.834 & 0.854 & 0.648 \\
        Sem Refine (SensatUrban) & 0.674 & 0.824 & 0.860 & 0.663 \\
        \hline
        2D Seg (STPLS3D) & 0.444 & 0.676 & 0.565 & 0.061 \\
        Sem Refine (STPLS3D) & 0.528 & 0.667 & 0.608 & 0.038 \\
        \hline
        2D Seg (STPLS3D finetune) & 0.652 & 0.827 & 0.838 & 0.638 \\
        Sem Refine (STPLS3D finetune) & 0.657 & 0.814 & 0.850 & 0.623 \\
        \hline
  \end{tabular}}
\end{table}

\begin{figure*}[t]
  \centering
  \includegraphics[width=0.78\linewidth,trim=0mm 0mm 0mm 0mm, clip]{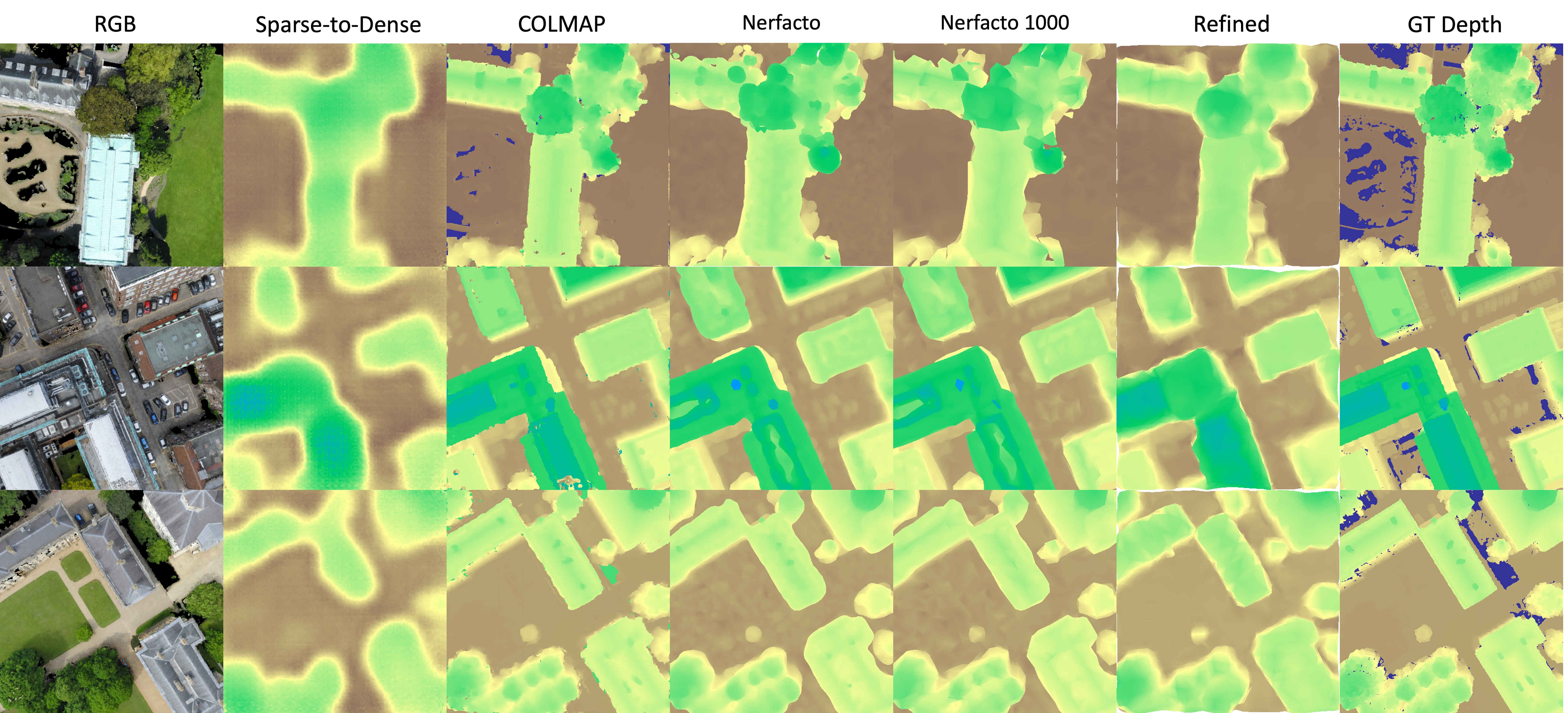}
  \caption{\reviewersix{Depth reconstruction comparison on the SensatUrban dataset among Sparse-to-Dense \cite{Ma2018Sparse}, COLMAP \cite{schoenberger2016mvs}, Nerfacto \cite{Tancik2023Nerfstudio}, Nerfacto restricted to 1000 mesh vertices, and our method.}}
  \label{fig:compare_others}
\end{figure*}

\begin{table*}[t]
  \caption{\reviewersix{Geometric error comparison among Sparse-to-Dense \cite{Ma2018Sparse}, COLMAP \cite{schoenberger2016mvs}, Nerfacto \cite{Tancik2023Nerfstudio}, and our method using metrics defined in \cite{Sun_2021_CVPR}. Nerfacto Mesh (1000) uses about 1000 mesh vertices, similar to our method. The other two mesh methods use about 20000 vertices.}}
  \label{tab:compare_others}
  \centering
  \resizebox{\linewidth}{!}{\begin{tabular}{|c|cccc|cccccc|}
        \hline
        \multirow{2}{*}{Method} & \multicolumn{4}{c|}{2D} & \multicolumn{6}{c|}{3D} \\
        \cline{2-11}
        & Abs Diff ($\ell_2$) $\downarrow$ & RMSE $\downarrow$ & Abs Rel $\downarrow$ & Sq Rel $\downarrow$ & Chamfer ($\ell_3$) $\downarrow$ & Accuracy $\downarrow$ & Completeness $\downarrow$ & Precision $\uparrow$ & Recall $\uparrow$ & F-score $\uparrow$\\
        \hline
        Initializaiton & 1.865 & 3.210 & 0.029 & 0.218 & 6.692 & 1.319 & 1.173 & 0.192 & 0.202 & 0.197 \\
        Refinement & 1.000 & 1.792 & 0.015 & 0.056 & 1.787 & 0.767 & 0.778 & 0.288 & 0.298 & 0.293 \\
        Semantic Refinement & 0.976 & 1.715 & 0.014 & 0.053 & 1.735 & 0.754 & 0.772 & 0.293 & 0.301 & 0.297 \\
        \hline
        Depth Completion - Sparse-to-Dense & 1.987 & 2.949 & 0.029 & 0.148 & 6.136 & 1.283 & 1.432 & 0.139 & 0.137 & 0.138\\
        MVS - COLMAP & 0.425 & 1.741 & 0.006 & 0.054 & 3.358 & 0.638 & 0.914 & 0.464 & 0.405 & 0.432 \\
        MVS - COLMAP Mesh & 0.588 & 1.612 & 0.008 & 0.039 & 3.379 & 0.980 & 0.724 & 0.320 & 0.361 & 0.339 \\
        NeRF - Nerfacto Mesh & 1.217 & 2.444 & 0.016 & 0.102 & 5.818 & 1.364 & 0.947 & 0.181 & 0.221 & 0.199\\
        NeRF - Nerfacto Mesh (1000) & 1.178 & 2.395 & 0.016 & 0.099 & 5.351 & 1.258 & 1.018 & 0.195 & 0.212 & 0.203 \\
        \hline
  \end{tabular}}
\end{table*}

\subsection{Comparison with Other Methods}
\label{sec:compare_others}
In this section, we compare our approach with depth completion, multi-view stereo, \reviewersix{and NeRF techniques} on the SensatUrban dataset \cite{Hu2021Sensat} with 1000 sparse depth measurements. The qualitative results are shown in Fig.~\ref{fig:compare_others}. The quantitative results are shown in Table~\ref{tab:compare_others}. \reviewersix{To evaluate the reconstruction quality of the different methods comprehensively, we report multiple 2D and 3D reconstruction accuracy metrics \cite{Sun_2021_CVPR}. The threshold distance for 3D precision and recall is set to 0.5m due to the large scale of the reconstructed mesh.}


\reviewera{
Depth completion methods take a sparse depth image and other inputs, such as an RGB image, and recover a dense depth image. While there are many depth completion algorithms, few focus on the aerial image domain. We compare against the Sparse-to-Dense method \cite{Ma2018Sparse}, an end-to-end deep learning regression model, because it considers a similar problem setting and uses similar feature extraction as our approach. The Sparse-to-Dense model consists of a ResNet feature extraction encoder and per-pixel depth regression decoder. We trained both our model and Sparse-to-Dense with a ResNet-34 feature extractor, thus focusing the comparison on the performance of the pure 2D learning and per-pixel depth regression of Sparse-to-Dense versus the joint 2D-3D learning for mesh refinement of our method. The results in Fig.~\ref{fig:compare_others} and Table~\ref{tab:compare_others} show that the Sparse-to-Dense model is not as accurate as our method on the SensatUrban dataset, and it is beneficial to utilize our joint 2D-3D learning technique. At least on this dataset, it is challenging for Sparse-to-Dense to regress an accurate dense depth map, while, using the same 2D feature extraction network, our mesh initialization and refinement method performs better at reconstructing the 3D scene.}



\reviewerb{
We also compare our method with COLMAP \cite{schoenberger2016mvs}, a multi-view stereo technique that recovers 3D structure from a series of calibrated images \reviewersix{using pixelwise view selection for depth and normal estimation}. In this comparison, we used ground-truth camera poses and skip the SfM step. For each frame, we manually select neighboring frames within a radius of $50$ m for multi-view stereo matching. We use the reconstructed dense depth images to obtain a global point cloud, generate a global mesh, and crop the mesh at each camera pose to obtain local meshes associated with each frame. For the meshing step, we compared Poisson reconstruction \cite{Kazhdan2006Poisson} and Delaunay mesh reconstruction \cite{Labatut2009Surface} and found that due to point cloud noise the Delaunay reconstruction performs better. The COLMAP is the dense depth estimation result, while the COLMAP Mesh is the subsequent meshing result. For COLMAP, we convert the dense depth images to a point cloud and sample 10000 points to compute the $\ell_3$ error. For COLMAP Mesh, we render the local mesh to get a rendered depth image to compute the $\ell_2$ error. The results are shown in Fig.~\ref{fig:compare_others} and Table~\ref{tab:compare_others}. The COLMAP method generally has better depth reconstruction measured by $\ell_2$ error, while our method has lower $\ell_3$ error because of the implicit regularization in our mesh reconstruction. The 3D error $\ell_3$ can be large when outliers appear in the reconstruction. COLMAP achieves better reconstruction at the cost of heavy computation for the MVS step. It takes around 4 seconds per frame to recover a dense depth image using GPU, while meshing requires additional time. Our method is much faster, with $0.07$ s per frame on a desktop with GeForce RTX 2080 Ti GPU and $0.45$ s per frame on a Jetson AGX Xavier edge computing platform. When it comes to online mesh reconstruction on a resource-constrained platform, our method offers an advantage over MVS.
}

\reviewersix{Finally, we compare with Nerfacto \cite{Tancik2023Nerfstudio}, a NeRF model that combines components from recent NeRF papers to achieve a balance between speed and quality. As a NeRF model, Nerfactor takes posed RGB images and constructs an implicit 3D scene represented by a deep neural network. We used ground-truth camera poses for each image and trained separate Nerfacto models for each test image sequences. One in ten frames was chosen as an evaluation image during training. Notice here no depth images are used for the NeRF training. We observed that depth images rendered directly from the trained Nerfacto model have inconsistent depth across frames. Therefore, we exported a mesh model using the Poisson surface reconstruction \cite{Kazhdan2006Poisson} implemented in Nerfstudio \cite{Tancik2023Nerfstudio}. Nerfacto exports meshes with different vertex density. A dense mesh has about 20000 vertices for each camera view, while a sparse mesh has about 1000, which is similar to our method's mesh vertex density. To compute the 2D metrics in Table~\ref{tab:compare_others}, we rendered depth images from the mesh. The evalution results for Nerfacto are shown in Fig.~\ref{fig:compare_others} and Table~\ref{tab:compare_others}. Qualitatively, Nerfacto has similar reconstruction accuracy with our method but the quantitative metrics indicate that it is not as good as our method. Furthermore, Nerfacto needs to be trained for each novel environment and the training can hardly meet real-time requirement, while our method can make the inference faster.
}


\subsection{Ablation Studies}
\label{sec:ablation}

Sec.~\ref{sec:exp_WHU} compared the effect of using RGB, rendered depth, and Euclidean distance transform as inputs for the mesh reconstruction model. This section reports additional ablation studies on the SensatUrban dataset. We use a 1024-vertex mesh model and 1000 sparse depth measurements for training and testing. We evaluate the effects of mesh initialization, types of 2D input data, and number of graph convolution stages on the geometric mesh reconstruction accuracy. We also evaluate the performance effect of joint metric-semantic training and the choice of a semantic loss function.


\subsubsection{Mesh Initialization}
An important aspect of our model in Sec.~\ref{sec:technical_approach} is the separation of the mesh initialization stage from the mesh refinement stage. The mesh initialization stage allows the data-driven refinement stage to focus on learning the mesh vertex deformation residuals instead of absolute vertex coordinates. To demonstrate the effectiveness of this design, we compare our model to a baseline model which applies the refinement stage directly to a flat initial mesh. The baseline model, Flat Init, deforms a flat initial mesh with vertex depth specified by the mean of the sparse depth measurements. Table \ref{tab:ablation} shows that the Flat Init model makes the 2D-3D learning problem challenging, and the model performs even worse than purely geometric initialization as in Sec.~\ref{sec:mesh_init}.


\subsubsection{2D Input Channels} In \eqref{eq:2d_inputs}, we concatenate an RGB image $\bfI$ (RGB), rendered depth $\rho_D(\mathcal{M}^{\text{int}})$ (RD) and a Euclidean distance transform $\bfE(\bfD^s)$ (EDT) to form a 5-channel input image used for 2D feature extraction. Table \ref{tab:ablation} evaluates the role of the different 2D inputs on the overall mesh reconstruction performance. The results indicate that the RGB information plays the most important role in refining the initialized mesh. The model RD+EDT that does not use RGB features performs the worst. Adding RD and EDT inputs to the RGB gives an additional boost to the accuracy. 


\subsubsection{Number of graph convolution stages} 
Table \ref{tab:ablation} also evaluates the effect of one (1 Stage) vs two (Our Model) graph convolution stages in the geometric mesh refinement (Sec.~\ref{sec:geo_mesh_refine}). The first GCN stage contributes the most to the geometric refinement, while the second GCN stage further refines the results.

\subsubsection{Semantic Loss Function}
Finally, we discuss the choice of a semantic loss function $\ell_{\bfS}$ in \eqref{eq:loss_sem_dice}. Instead of the Dice loss in \eqref{eq:loss_sem_dice}, three other semantic loss functions may be considered.
\begin{itemize}
    \item The cross entropy loss is widely used for semantic segmentation. Given two stochastic vectors $\bfalpha,\bfbeta \in [0,1]^s$, the cross entropy loss is defined as:
\begin{equation}
\label{eq:loss_sem_entropy}
\begin{aligned} 
\text{CE}(\bfalpha,\bfbeta) & = -\sum_{i=1}^s \bfbeta_i \log(\bfalpha_i), \\
\ell_{\bfS1}(\calM,\bfS) &:= \text{mean}( \text{CE}(\sigma(\rho_{S}(\calM)),\bfS)),    
\end{aligned}
\end{equation}
where $\text{CE}$ is applied to the elements $\sigma(\rho_{S,ij}(\calM)) \in [0,1]^s$ and $\bfS_{ij} \in [0,1]^s$ of the tensors of predicted and ground-truth semantic class probabilities.
    \item The focal loss \cite{Lin2020Focal} is a variation of cross entropy, focusing on hard misclassified examples:
\begin{equation}
\label{eq:loss_sem_focal}
\begin{aligned} 
\text{FL}(\bfalpha,\bfbeta) & = -\sum_i \bfbeta_i (1-\bfalpha_i) \log(\bfalpha_i), \\
\ell_{\bfS2}(\calM,\bfS) &:= \text{mean}( \text{FL}(\sigma(\rho_{S}(\calM)),\bfS)).
\end{aligned}
\end{equation}
    \item The Jaccard loss \cite{Jaccard1912} measures the negative Intersection over Union (IoU) between the ground-truth and predicted semantic segmentation:
\begin{equation}
\label{eq:loss_sem_jaccard}
\ell_{\bfS3}(\calM,\bfS) := - \frac{|\sigma(\rho_{S}(\calM)) \cdot \bfS|}{|\sigma(\rho_{S}(\calM))| + |\bfS| - |\sigma(\rho_{S}(\calM)) \cdot \bfS|},
\end{equation}
where, as in \eqref{eq:loss_sem_dice}, $|\cdot|$ sums up all the absolute values of the elements.
\end{itemize}
In Table \ref{tab:sem_sensat}, we see that the Jaccard loss in \eqref{eq:loss_sem_jaccard} leads to good segmentation performance, outperforming the Dice loss in \eqref{eq:loss_sem_dice} for some categories. The Cross Entropy and the Focal losses are not as good.
In Table \ref{tab:geo_sensat}, we see that the Cross Entropy and the Jaccard loss both outperform the Focal loss when considering their effect on the geometric reconstruction accuracy. The Dice loss leads to the best geometric reconstruction accuracy. Considering the joint geometric and semantic performance, we elected to use the Dice loss for our final model.

\begin{table}[t]
  \caption{\reviewer{Ablation study on geometric reconstruction error for geometric models. The definitions of the different models can be found in Sec. \ref{sec:ablation}.}}
  \label{tab:ablation}
  \centering
  \begin{tabular}{|c|cc|}
        \hline
        Method & Depth ($\ell_2$) & Chamfer ($\ell_3$) \\
        \hline
        Geo Init & 1.865 & 6.725 \\
        \hline
        Flat Init & 4.646 & 20.655 \\
        \hline
        RD+EDT & 1.761 & 5.791 \\
        RGB & 1.521 & 3.750 \\
        RGB+RD & 1.070 & 2.138 \\
        \hline
        1 Stage & 1.015 & 1.828 \\
        \hline
        Our Model & \textbf{1.000} & \textbf{1.815} \\
        \hline
  \end{tabular}
\end{table}

\subsection{Memory and Computation Complexity}
\label{sec:complexity}

The reconstructed mesh model is a more efficient representation than a dense depth image. A dense depth image requires $512\times 512 \approx 0.26M$ parameters, and a semantic image also requires the same number of parameters. Our mesh model with fixed face topology only needs storage of the 3D vertex coordinates and the semantic labels. With 1024 vertices, our semantic mesh model requires only $2\%$ of the depth and semantic image parameters to obtain a high-fidelity reconstruction of a camera view. 
\reviewersix{Our model has about 21M parameters (ResNet and GCN) and takes about 3GB GPU memory during inference.}
\reviewera{We report the inference time of our model on different computation platforms in Table~\ref{tab:speed}. The results show that our mesh reconstruction algorithm can achieve 2 Hz on an embedded NVIDIA Jetson AGX Xavier computer, making it applicable for real-time deployment onboard a robot system. Regarding timing the baseline algorithm for sparse-depth triangulation, we evaluated its run-time frequency to be at around 20 Hz on the same platform.}

\begin{table}[t]
  \caption{\reviewera{Prediction time (second) on different NVIDIA devices.}}
  \label{tab:speed}
  \centering
  \begin{tabular}{|c|ccc|}
        \hline
        Platform & Initialization & Refinement & Total \\
        \hline
        Jetson AGX Xavier w/o GPU & 0.45 & 0.75 & 1.20 \\
        Jetson AGX Xavier GPU & 0.30 & 0.15 & 0.45 \\
        GeForce RTX 2080 Ti GPU & 0.05 & 0.02 & 0.07 \\
        \hline
  \end{tabular}
\end{table}

\subsection{Limitations}
\label{sec:limitation}
\reviewersix{
Our 3D metric-semantic mesh reconstruction algorithm can run efficiently on an embedded computer but as a result the number mesh vertices used for reconstruction is limited, which in turn affects the geometric reconstruction accuracy. Furthermore, local meshes are generated using only a single camera frame without multi-view constraints, making it challenging to achieve consistent mesh merging into a global model. Potential avenues for future work that may improve the reconstruction quality include adaptively increasing the mesh vertices depending on the image feature distribution, considering techniques like deformable convolution \cite{Dai_2017_ICCV} for associating the 3D mesh vertices with the 2D image features, utilizing sparse depth measurement uncertainty (e.g., keypoint covariances provided by SLAM) for weighted interpolation during the image features to vertex association, and improving the global mesh merging approach with multi-view constraints. 
}

\section{Conclusion}
\label{sec:conclusion}

This work introduces an approach for 3D metric-semantic mesh reconstruction from RGB image and sparse depth measurements. Compared to methods that utilize only sparse depth for mesh initialization or triangulation, our approach provides more accurate geometric reconstruction by utilizing RGB image features. Compared to 2D semantic segmentation methods, our semantic reconstruction eliminates classification inaccuracies by inferring an underlying 3D mesh structure. The joint metric-semantic reconstruction approach improve geometric accuracy further by utilizing semantic information and provides memory savings compared to dense image depth and segmentation techniques. Employing our method in combination with feature- and keyframe-based odometry techniques allows reconstruction of global dense metric-semantic mesh models with utility in environmental monitoring and semantic navigation applications.

{\small
\bibliographystyle{cls/IEEEtran}
\bibliography{bib/ref.bib}
}

%


\begin{IEEEbiography}[{\includegraphics[width=1in,height=1.25in,clip,keepaspectratio]{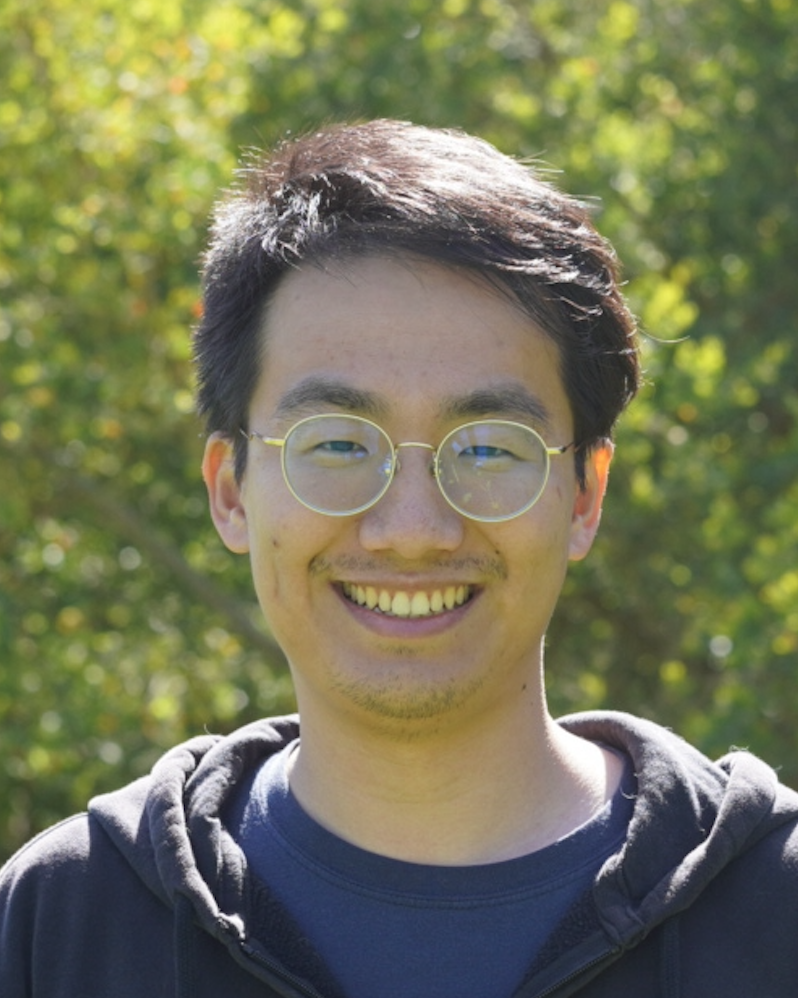}}]{Qiaojun Feng} 
(S'19) is a Ph.D. Candidate of Electrical and Computer Engineering at the University of California San Diego, La Jolla, CA, USA. He obtained a B.Eng. in Automation from Tsinghua University, Beijing, China in 2017 and an M.S. in Electrical Engineering from the University of California San Diego in 2019. His research focuses on mobile robot autonomy, and particularly on metric-semantic perception and mapping.
\end{IEEEbiography}


\begin{IEEEbiography}[{\includegraphics[width=1in,height=1.25in,clip,keepaspectratio]{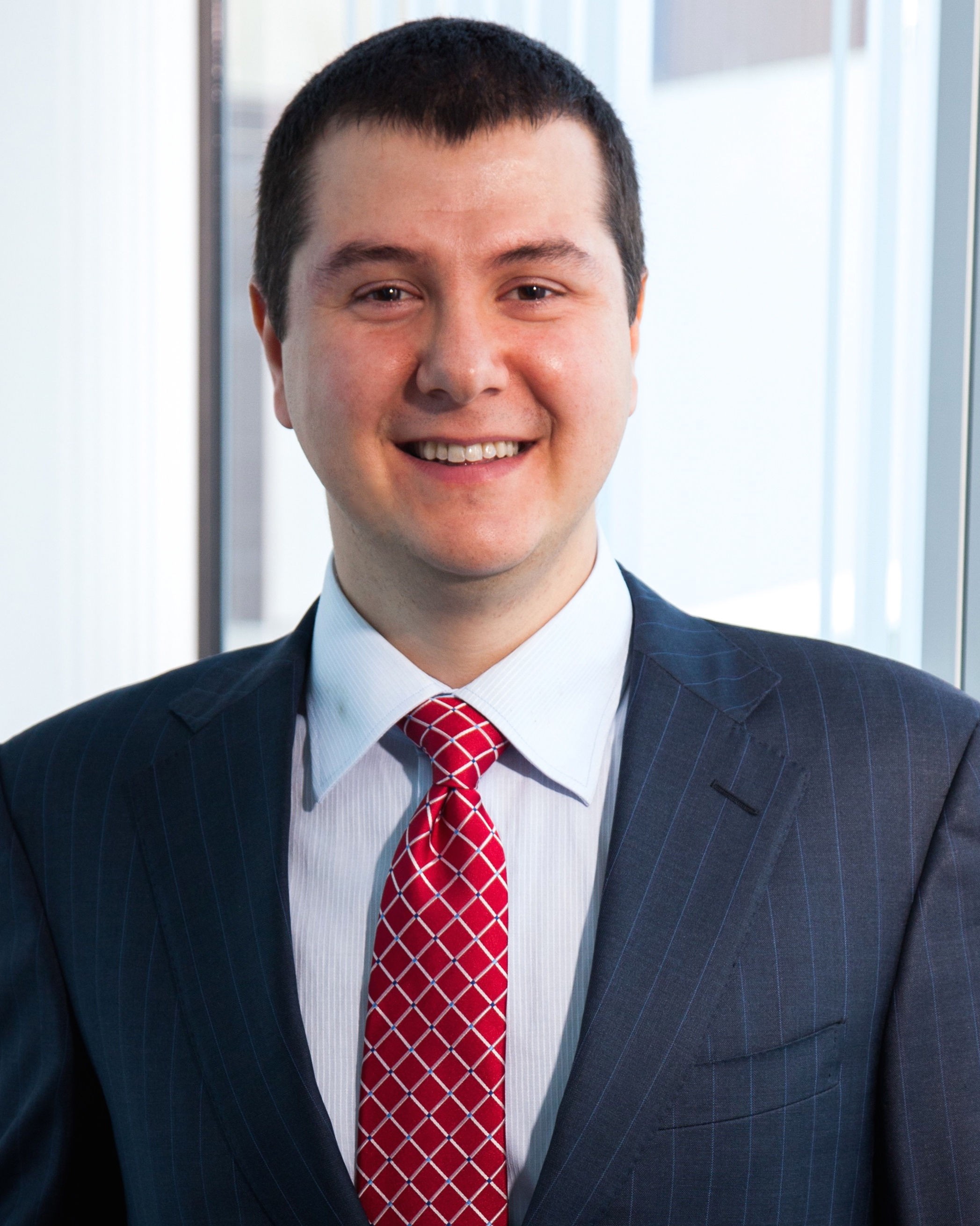}}]{Nikolay Atanasov}
(S'07-M'16-SM'23) is an Assistant Professor of Electrical and Computer Engineering at the University of California San Diego, La Jolla, CA, USA. He obtained a B.S. degree in Electrical Engineering from Trinity College, Hartford, CT, USA in 2008 and M.S. and Ph.D. degrees in Electrical and Systems Engineering from the University of Pennsylvania, Philadelphia, PA, USA in 2012 and 2015, respectively. Dr. Atanasov's research focuses on robotics, control theory, and machine learning, applied to active perception problems for autonomous mobile robots. He works on probabilistic models that unify geometric and semantic information in simultaneous localization and mapping (SLAM) and on optimal control and reinforcement learning algorithms for minimizing probabilistic model uncertainty. Dr. Atanasov's work has been recognized by the Joseph and Rosaline Wolf award for the best Ph.D. dissertation in Electrical and Systems Engineering at the University of Pennsylvania in 2015, the Best Conference Paper Award at the IEEE International Conference on Robotics and Automation (ICRA) in 2017, the NSF CAREER Award in 2021, and the IEEE RAS Early Academic Career Award in Robotics and Automation in 2023.
\end{IEEEbiography}




\end{document}